\PassOptionsToPackage{table,xcdraw}{xcolor}
 \documentclass[final,3p,times]{elsarticle}
\usepackage{amssymb}
\usepackage{tikz}
\usepackage{pgfplots}
\usepackage{sansmath}
\pgfplotsset{compat=1.18}
\usepackage{booktabs}
\usepackage{multirow}
\usepackage{colortbl}
\usepackage[utf8]{inputenc}
\usepackage{subcaption}
\DeclareCaptionJustification{slightlyright}{\hspace{2cm}} % define new style

\usepackage{amsmath}
\usepackage{mathrsfs}
\usepackage{xcolor}
\usepackage[most]{tcolorbox}
\usepackage{lipsum} % 示例文本用，可删
\usepackage{lineno}

\tcbset{
  promptstyle/.style={
    colback=gray!10,    % 背景颜色
    colframe=gray!70,   % 边框颜色
    fonttitle=\bfseries,
    boxrule=0.5pt,
    arc=2mm,            % 圆角
    left=2mm,
    right=2mm,
    top=1mm,
    bottom=1mm,
    enhanced,
    breakable           % 可以跨页
  }
}

\newcommand{\reals}{\mathbb{R}}
\journal{Information Fusion}

\makeatletter
\def\ps@pprintTitle{%
 \let\@oddhead\@empty
 \let\@evenhead\@empty
 \let\@oddfoot\@empty
 \let\@evenfoot\@empty
}
\makeatother

\begin{document}

\begin{frontmatter}



\title{Information-Theoretic Graph Fusion with Vision-Language-Action Model for Policy Reasoning and Dual Robotic Control}

% \affiliation[label1]{organization={},
%%\author[addr_a]{Shunlei Li\fnref{equal1}\corref{cor1}}
%%\author[addr_b]{Longsen Gao\fnref{equal1}}

\author[addr_a]{Shunlei Li\corref{cor1}}
\author[addr_b]{Longsen Gao}
\author[addr_c]{Jin Wang}
\author[addr_d]{Chang Che}
\author[addr_e]{Xi Xiao}
\author[addr_a]{Jiuwen Cao}
\author[addr_f]{Yingbai Hu\corref{cor1}}
\author[addr_g]{Hamid Reza Karimi}

\cortext[cor1]{Corresponding author, Email: Shunlei Li, shunlei.li@outlook.com; Yingbai Hu, yingbai.hu@tum.de}

\address[addr_a]{Machine Learning and I-health International Cooperation Base of Zhejiang Province,\\
Artificial Intelligence Institute, Hangzhou Dianzi University, Zhejiang, China, 310018}

\address[addr_b]{Electrical and Computer Engineering Department, University of New Mexico, Albuquerque, United States, 87106}

\address[addr_c]{Dynamic Robot Systems Group, Oxford Robotics Institute, University of Oxford, United Kingdom, OX26NN}

\address[addr_d]{Mechanical and Aerospace Engineering Department, The George Washington University, DC, United States, 22202}

\address[addr_e]{Department of Computer Science, University of Alabama at Birmingham, Alabama, United States, 35294}

\address[addr_f]{The School of Computation, Information and Technology, Technical University of Munich, Germany, 85748}

\address[addr_g]{Department of Mechanical Engineering, Politecnico di Milano, Milan, Italy, 20156}

%% Abstract
\begin{abstract}
Teaching robots dexterous skills from human videos remains challenging due to the reliance on low-level trajectory imitation, which fails to generalize across object types, spatial layouts, and manipulator configurations. We propose Graph-Fused Vision–Language–Action (GF-VLA), a framework that enables dual-arm robotic systems to perform task-level reasoning and execution directly from RGB(-D) human demonstrations.
GF-VLA first extracts Shannon-information-based cues to identify hands and objects with the highest task relevance, then encodes these cues into temporally ordered scene graphs that capture both hand–object and object–object interactions. These graphs are fused with a language-conditioned transformer that generates hierarchical behavior trees and interpretable Cartesian motion commands. To improve execution efficiency in bimanual settings, we further introduce a cross-hand selection policy that infers optimal gripper assignment without explicit geometric reasoning.
We evaluate GF-VLA on four structured dual-arm block assembly tasks involving symbolic shape construction and spatial generalization. Experimental results show that the information-theoretic scene representation achieves over 95\% graph accuracy and 93\% subtask segmentation, supporting the LLM planner in generating reliable and human-readable task policies. When executed by the dual-arm robot, these policies yield 94\% grasp success, 89\% placement accuracy, and 90\% overall task success across stacking, letter-building, and geometric reconfiguration scenarios, demonstrating strong generalization and robustness across diverse spatial and semantic variations.

\end{abstract}

%% Keywords

\end{frontmatter}


\section{Introduction}

Developing robotic systems capable of robust manipulation in unstructured and dynamic environments is a central challenge in embodied artificial intelligence. Effective manipulation requires robots to perceive complex scenes, explicitly reason about physical interactions, and reliably execute multi-step tasks under uncertainty. Traditional vision-based methods, such as visual servoing~\cite{chaumette2016visual}, typically rely on monocular or RGB-D inputs and fixed sensor calibrations~\cite{piacenza2024vfas}, imposing rigid assumptions that rarely hold in real-world scenarios. Such approaches~\cite{manderson2020learning, gao2016deep, cao2021six, wang2024hypermotion} often exhibit limited robustness due to susceptibility to sensor noise, occlusions, and physical contacts~\cite{yuan2021multi}, and they commonly neglect complementary sensing modalities, including tactile feedback and proprioception. Consequently, their ability to generalize beyond highly controlled environments remains substantially limited.

Recently, Vision-Language-Action (VLA) models have become increasingly prominent for addressing limitations inherent in traditional robotic manipulation approaches~\cite{brohan2023rt2, kim2024openvla, black2024pi0,zhang2025invertible,li2024robonursevlaroboticscrubnurse}. 
By leveraging large-scale multimodal pretraining on extensive image-text datasets, these models effectively translate natural language instructions into executable actions. The integration of powerful pretrained vision-language models enables robots to comprehend semantically diverse instructions and perform tasks in an open-ended manner, significantly expanding their applicability in novel scenarios. Moreover, VLAs inherently support semantic-level reasoning, allowing robots to interpret abstract goals rather than relying purely on low-level perceptual cues. Despite these compelling advantages, current VLA frameworks exhibit substantial limitations, especially when precise manipulation or physical interaction is required~\cite{choi2024handnerf,shridhar2022cliport, jang2022bc, mendonca2023rtx}. A primary issue is their limited ability to model structured, dynamic physical relationships explicitly; current VLAs typically lack inductive biases necessary to capture fine-grained interactions between objects and agents over time. Consequently, they frequently struggle to generate physically plausible execution plans, particularly under ambiguous instructions, unseen object arrangements, or conditions involving physical contacts and precise temporal coordination~\cite{nair2022bct, huang2023inner,hu2024fusion}. Furthermore, their performance heavily relies on the scale and distribution of the pretraining datasets, making them vulnerable to distributional shifts encountered in practical robotic tasks.

Despite the compelling advantages of VLA models, current frameworks exhibit substantial limitations when precise manipulation or physical interaction is required~\cite{stepputtis2020language}. A primary issue is their limited ability to model structured dynamic physical relationships explicitly, as they typically lack the inductive biases necessary to capture fine-grained interactions between objects and agents over time~\cite{xie2020deep}. Consequently, they frequently struggle to generate physically plausible execution plans, particularly under ambiguous instructions or conditions involving physical contact. Furthermore, while policy learning approaches effectively acquire skills in structured environments, they often suffer from poor generalization when deployed in novel contexts.

In parallel, several alternative methodologies have been extensively explored to address some of these limitations. Policy learning approaches, including reinforcement learning (RL) and imitation learning, enable robots to acquire manipulation behaviors directly from interaction data or demonstrations~\cite{li2018deep, rajeswaran2018learning, kalashnikov2018qtopt, yeh2023ilovi,mao2025liquid}. While effective at learning sophisticated skills in structured environments, these methods typically require vast amounts of training data, making them computationally expensive and sample-inefficient. They also suffer from poor generalization when deployed in novel contexts due to overfitting on specific task configurations. Information-theoretic methods provide a principled mathematical framework for systematically quantifying uncertainty and identifying crucial interactions within multimodal data streams~\cite{goyal2019infobot, hu2024efficient,zhou2024maxmi, merlo2025information}. However, despite their theoretical rigor, these methods are challenging to scale to complex manipulation tasks and rarely integrate effectively with high-level semantic task reasoning or instruction interpretation. Additionally, transfer learning methodologies have been proposed to enhance the generalization capabilities of robotic policies across different domains or task variations~\cite{bousmalis2018using, peng2020learning, jaquier2025transfer}. Yet, these approaches frequently depend on extensive target-domain data and task-specific fine-tuning, limiting their scalability and interpretability. Consequently, none of these existing paradigms alone sufficiently address the fundamental challenge of robustly unifying structured physical interaction modeling with semantic-level task reasoning, which is crucial for achieving flexible and reliable robotic manipulation in realistic environments.

\begin{figure*}[!t]
    \centering
    \includegraphics[width=0.95\linewidth]{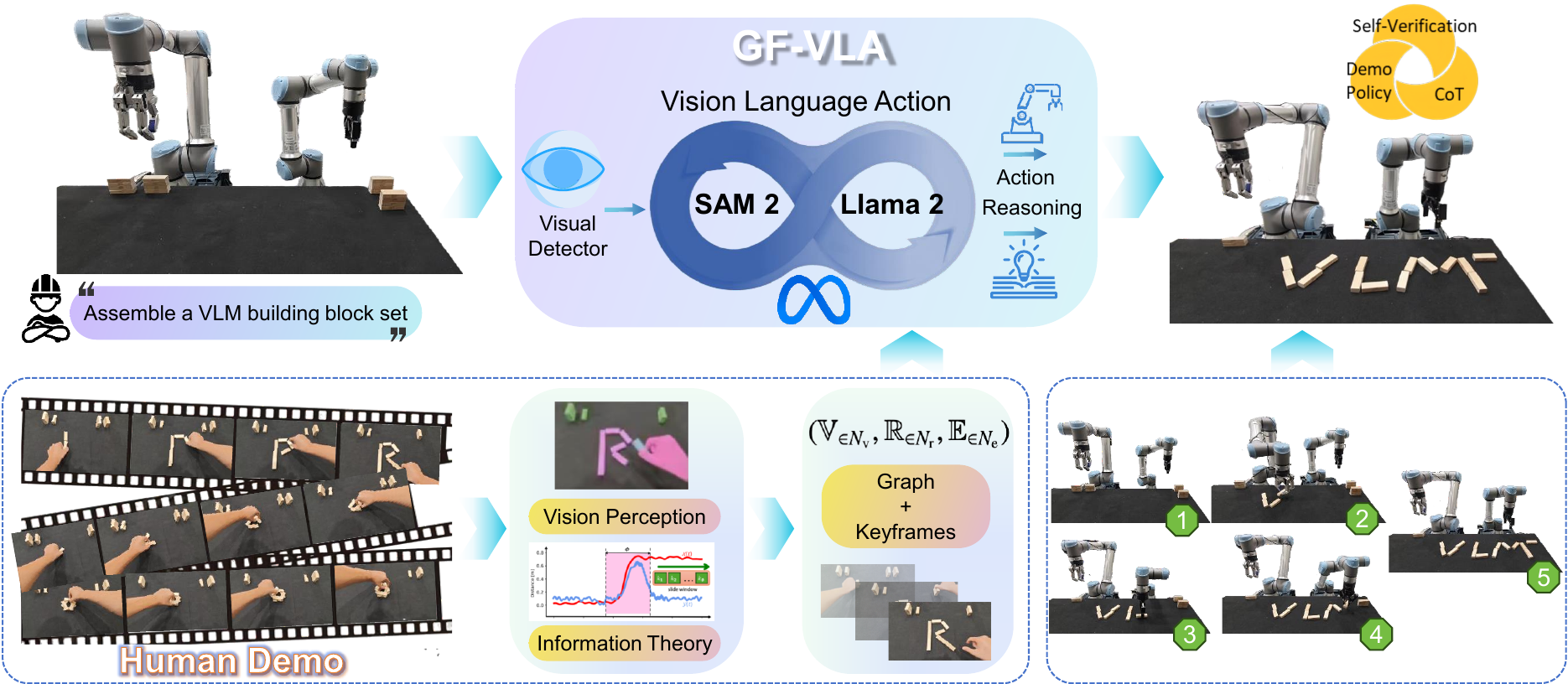}
    \caption{An overview of the GF-VLA framework that performs policy transfer from a single human demonstration to a dual-arm robot manipulation task.
    }
    \label{fig:overview_gf_vla}
\end{figure*}

The central motivation of this work, therefore, stems from the critical need to bridge the gap between high-level semantic reasoning and low-level physical understanding~\cite{azzolini2025cosmos}. We argue that achieving truly robust and generalizable robotic manipulation requires a unified framework that not only interprets abstract language goals but also grounds them in a structured, information-theoretic representation of the physical scene. By explicitly fusing these modalities, we aim to endow robots with the capability to reason about task dynamics and execute complex bimanual interactions with the adaptability seen in human demonstrations. To address these challenges, we advocate for a framework as shown in Figure \ref{fig:overview_gf_vla} that explicitly integrates structured physical interaction modeling with multimodal semantic reasoning. Such a framework must leverage structured inductive biases to enable robust execution, flexible generalization, and interpretable action planning in dynamic environments. 

In this paper, we propose Graph-Fused VLA (GF-VLA), a unified framework combining structured scene graph representations derived from multimodal human demonstrations with vision-language-action reasoning. Our approach constructs temporally structured scene graphs from multimodal demonstration data—including video sequences, and object trajectories by leveraging entropy and mutual information to systematically encode dynamic physical interactions. These interaction-aware graphs are then integrated into VLA model, enabling semantically grounded task planning and interpretable action generation. Moreover, we incorporate Chain-of-Thought (CoT) and self-verification prompting to facilitate explicit subgoal decomposition and enhance transparency of robotic execution. Our method further leverages transfer learning from limited demonstration data, significantly improving generalization to unseen objects and tasks.

Our main contributions are summarized as follows:
\begin{itemize}
    \item We revisit and significantly extend the concept of scene representation in robotic manipulation by proposing a principled information-theoretic approach to construct temporally and spatially structured scene graphs from multimodal human demonstration data with semantic segmentation. Our approach explicitly encodes dynamic physical interactions, addressing a critical gap left unfilled by existing perceptual frameworks.

    \item We establish a novel unified paradigm, GF-VLA, that systematically integrates structured physical interaction modeling with vision-language-action reasoning for the first time. This framework fundamentally advances structured semantic reasoning in robotics, enabling robust, interpretable, and generalizable manipulation behaviors in dynamic environments.

    \item We substantially enhance interpretability and transparency in robotic planning by pioneering the integration of Chain-of-Thought prompting with vision-language-action models. This innovation provides explicit and structured subgoal decomposition, significantly improving policy understandability and execution fidelity.

    \item Through rigorous empirical evaluation, we set a new state-of-the-art benchmark across challenging dual-arm manipulation scenarios, demonstrating our method’s superior robustness, generalization, and interpretability compared to leading contemporary approaches. Our results clearly indicate that GF-VLA represents a major step forward toward practical, robust, and trustworthy embodied intelligence.
\end{itemize}

\section{Related works}

%%1. vla方法介绍，并列举相对突出的工作 如openvla等
%%2. information theory方法介绍 并列举相对应的论文 如tro paper:Exploiting Information Theory for Intuitive Robot和 https://arxiv.org/abs/2407.15086

\subsection{Visual-Language-Action models}
VLA models aim to learn end‑to‑end
mappings from multimodal inputs to robot control commands and
have become a central research theme in robotic learning.  Current
efforts concentrate on refining \emph{action‑generation strategies}
while enhancing representational power and generalisation.
OpenVLA\cite{kim2024openvla}, the first open‑source VLA framework, adopts RT‑2’s\cite{zitkovich2023rt}
autoregressive language‑model paradigm and encodes actions as
sequences of discrete tokens, harnessing large language models to
directly control robots from natural‑language commands.  Yet
discretisation inevitably compromises the fidelity of continuous
motions, limiting suitability for high‑frequency closed‑loop control.
Octo\cite{mees2024octo} alleviates this bottleneck by introducing a diffusion model that
probabilistically represents continuous actions, inspiring later
flow‑matching approaches, though its training efficiency remains low.
HybridVLA\cite{liu2025hybridvla} and Diffusion‑VLA\cite{wen2025diffusionvla} combine diffusion with autoregressive
generation to accelerate both training and inference, while CogACT\cite{li2024cogact}
replaces OpenVLA’s token predictor with a diffusion transformer,
achieving performance close to $\pi_0$.
The $\pi_0$\cite{DBLP:journals/corr/abs-2410-24164} model extends the \textit{Transfusion} framework with a
mixture‑of‑experts design: a vision--language expert handles perception
and semantics, whereas an action expert predicts high‑dimensional
continuous motions.  Conditional flow matching and \emph{action
chunking} precisely model continuous action distributions, boosting
task success rates and cross‑task generalisation, albeit at
substantial data and computational cost.  To improve efficiency,
$\pi_{0}$‑FAST\cite{DBLP:journals/corr/abs-2410-24164} introduces a novel tokenizer that
pre‑compresses motion segments via byte‑pair‑encoding–style rules,
markedly reducing training and inference latency on dexterous‑
manipulation datasets, though its autoregressive decoding is still
slower than flow‑matching generation.  The latest $\pi_{0.5}$\cite{DBLP:journals/corr/abs-2504-16054} model
fuses CoT‑VLA\cite{DBLP:conf/cvpr/ZhaoLKFZWLMHFHL25}’s chain‑of‑thought paradigm with hierarchical
task decomposition—first parsing high‑level commands into structured
sub‑tasks, then refining them into concrete motions—greatly enhancing
zero‑shot generalisation in open household environments.
LAPA\cite{DBLP:conf/iclr/YeJJJYPMTCLLL0Z25} learns from internet‑scale videos lacking robot‑action
labels, expanding available training sources.  RoboTwin\cite{DBLP:journals/corr/abs-2504-13059}
provides a real‑to‑simulation pipeline that leverages 3D generative
foundation models to produce diverse object meshes closely aligned
with real‑world counterparts, narrowing the sim‑to‑real gap, though
its reliance on predefined key‑points constrains tool‑interaction
modes.
Despite the significant progress made by VLA models in the field of robotic control, the cross-task generalization ability of models still needs improvement in open environments, requiring the establishment of more adaptive knowledge transfer mechanisms. 

\subsection{Information Theory}

Information Theory (IT) \cite{shannon1948mathematical} was originally established to quantify the inherent uncertainty within a signal and thereby characterize the stochastic nature of its potential states. This quantification is primarily accomplished through the computation of entropy over a defined probability distribution. Within the landscape of modern machine learning, information theory offers a rigorous theoretical basis for interpreting learning dynamics and formulating principles for regularization and generalization. Contemporary research continues to extrapolate these foundational concepts to tackle complex problems in neural network analysis \cite{anusha2024cloud}, multi-agent coordination \cite{he2025llm}, and distributed learning frameworks \cite{xu2024decentralized}. In the specific domain of robotic manipulation, researchers have leveraged these principles to bridge the gap between physical data and semantic understanding. For instance, the MaxMI framework \cite{zhou2024maxmi} introduces a Maximal Mutual Information criterion to precisely localize physical states that correspond to human-level semantics, which effectively mitigates the reliance on manual supervision or heuristic constraints. Furthermore, \cite{merlo2025exploiting} investigates the extraction of active scene elements to quantify the information transfer between the robotic hand and the manipulated object, marking a pioneering application of Shannon’s Information Theory to the analysis of manual tasks. Beyond analysis, information-theoretic objectives have been seamlessly integrated into policy learning architectures \cite{peng2024unsupervised,chen2025shieldagent}. By maximizing the information gain regarding action outcomes or prospective states, these methods significantly bolster the exploration efficiency of agents in uncertain environments. Such approaches prove particularly advantageous in contact-rich operational scenarios where sensory feedback is frequently obscured by noise and difficult to model with high precision.

In contrast to prior works that primarily emphasize action representation formats or generative efficiency, our proposed GF-VLA framework introduces a structured knowledge layer derived from human demonstrations in the form of information-theoretic behavior graphs. These graphs serve as intermediate symbolic representations that encode the temporal and semantic structure of task dynamics. By integrating them with visual observations and natural-language instructions, GF-VLA enables interpretable policy generation through an LLM head enhanced with chain-of-thought reasoning. This design improves policy generalization via symbolic abstraction and also supports real-time execution with local replanning, addressing both adaptability and robustness in dynamic multi-step dual-arm manipulation tasks.

\section{Information-Theoretic Scene Graphs}
\label{sec::scene_rep}
This section describes the detailed methodology for processing individual frames within a task demonstration, emphasizing the integration of entropy analysis applied to video clip sequences to effectively represent tasks through temporal scene graph structures. Initially, a perception module is responsible for the identification and localization of hands and objects in each video frame $k$ observed at a specific time instant. To acquire detailed interaction data, we employ the SAM2 model along with hand pose estimation~\cite{tewolde2025fingerposenet} that the module captures the hand pose information $\mathbf{q}^h_i(t) = \left[\mathbf{p}^h_i \ \  \boldsymbol{\theta}^h_i \right]^\top \in \mathbb{R}^6$ of $i_\mathrm{th}$ hand element in $\mathbb{H}_{\in N_h}$ set with $N_h \in \reals_+$ number of elements which $\mathbf{p}^h_i = [x^h_i \ \ y^h_i \ \ z^h_i ]^\top \in \mathbb{R}^3$ and $\boldsymbol{\theta}^h_i = [\alpha^h_i  \ \ \beta^h_i \ \ \gamma^h_i ]^\top \in \mathbb{R}^3$ denote the position and orientation of the hand, respectively, and the corresponding pose $\mathbf{q}_j^b(t) = \left[\mathbf{p}_j^b \ \ \boldsymbol{\theta}_j^b \right]^\top \in \mathbb{R}^6$ for $j_{\mathrm{th}}$ object in object set $o_m \in \mathbb{O}_{\in N_o}$ with $N_o \in \reals_+$ number of elements in which  $\mathbf{p}_j^b = [x_j^b  \ \ y_j^b \ \ z_j^b ]^\top \in \mathbb{R}^3$ and $\boldsymbol{\theta}_j^b = [\alpha_j^b  \ \ \beta_j^b \ \ \gamma_j^b ]^\top \in \mathbb{R}^3$ denote the position and orientation of the $j_{\mathrm{th}}$ object, respectively, within the observed scene. The positional data are then forwarded to an entropy calculation module, where an evaluation is performed on the informational content derived from each positional measurement, as well as the information exchanged between pairs of scene elements within a defined temporal window $\phi$ centered around $t$. Additionally, these positional signals undergo further processing to compute the mean spatial distances between hands and objects within the same temporal window $\phi$. The resulting processed data enables the construction of a structured representation known as a scene graph  $\texttt{SR}[k]$, effectively capturing relational dynamics among the detected scene elements. In this representation, graph nodes correspond to detected hands and objects at frame~$k$, with the graph representation capturing a set of relationships between nodes; edges indicate inferred interactions between connected pairs.
Finally, these generated graph structures are systematically stored for subsequent analysis. 
% Above revised done
\subsection{Information Theory in Robotic Manipulation}
To effectively represent tasks, accurately identifying the \textit{active} portion of a scene, defined as regions exhibiting significant dynamical variations is critical. By focusing explicitly on these active areas, we achieve a more concise and meaningful task representation that isolates the elements directly relevant to task execution and excludes non-essential on-table components. In this context, information theory (IT), initially formulated by Shannon~\cite{shannon1948mathematical}, provides a robust analytical framework. IT quantitatively assesses the informational content of signals independently of their semantic interpretations. The principal metric within IT, known as \emph{entropy}, quantifies the average uncertainty embedded in the signal, effectively characterizing the unpredictability or surprise associated with its possible states. Given a random variable $\mathcal{X}$ linked to an event space $\mathbb{X}_{\phi \in N_\mathrm{x}}$, the Shannon's entropy of $\mathbb{X}_{\in N_\mathrm{x}}$ is given by:
\begin{equation}
\label{eq:entropy}
    \mathcal{H}^\mathcal{X}(p) = -\epsilon \sum_{i=1}^{N_\mathrm{x}} p(x_i) \cdot \ln {p(x_i)},\ \epsilon \in \mathbb{R}_{+}
\end{equation}
where $\mathbb{X}_{\phi \in N_\mathrm{x}}:= \{x_1, \dots, x_{N_\mathrm{x}}\} = \{x{\scriptstyle (t_{\phi/2}-\tfrac{\phi}{2})}, \dots, x{\scriptstyle (t_{\phi/2}+\tfrac{\phi}{2})}\}$ in which $\phi$ denotes a sliding temporal window that we take to calculate entropy value $\mathcal{H}^{\mathcal{X}(t)}$ and $t$ denotes the time step within the temporal sliding window $\phi$; $x_i \in \mathbb{X}_{\in N_\mathrm{x}}$ is a discrete element from set $\mathbb{X}_{\in N_\mathrm{x}}$, and $\mathcal{X}:= x_i \sim \mathbb{X}_{\phi \in N_\mathrm{x}}$ denotes a random variable extracted from the set $\mathbb{X}_{\phi \in N_\mathrm{x}}$; $p(x_i)$ denotes the probability at $i_{\mathrm{th}}$ element in $\mathbb{X}_{\in N_\mathrm{x}}$, and $\epsilon \in (0,1)$ is a constant value. $\mathcal{H}^{\mathcal{X}(t_{\phi/2})}$ is the resulting entropy value, which is centered at the sliding window $\phi$. Note that Shannon's entropy is measured in bits. The higher the entropy value, the more bits are required to transmit the information contained in the signal. 

To provide a theoretical guarantee for using entropy as a motion indicator, we establish the following proposition regarding the lower bound of the entropy metric static conditions. For a position signal segment $\mathbb{X}_{\phi}$ within a temporal window $\phi$, the Shannon entropy $\mathcal{H}^X$ is minimized if and only if the object remains static relative to the quantization resolution $\xi$.

\begin{figure}[!t]
    \centering
    % \captionsetup[subfigure]{justification=centering, singlelinecheck=false, margin={1.20in, 6cm}} 
    \captionsetup{font=footnotesize}
    \begin{subfigure}[c]{0.48\textwidth}
        \centering
        \captionsetup{font=footnotesize,,margin={0.6cm,0cm}}\includegraphics[width=0.9\linewidth]{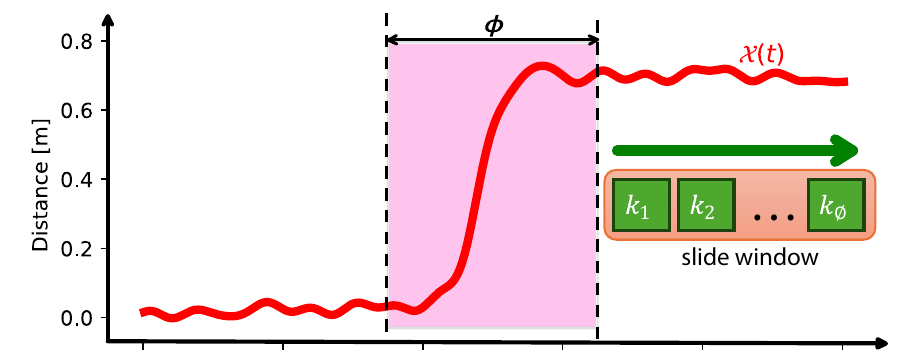}
        \subcaption{}
        \label{fig1::sub_1} 
        \includegraphics[width=0.9\linewidth]{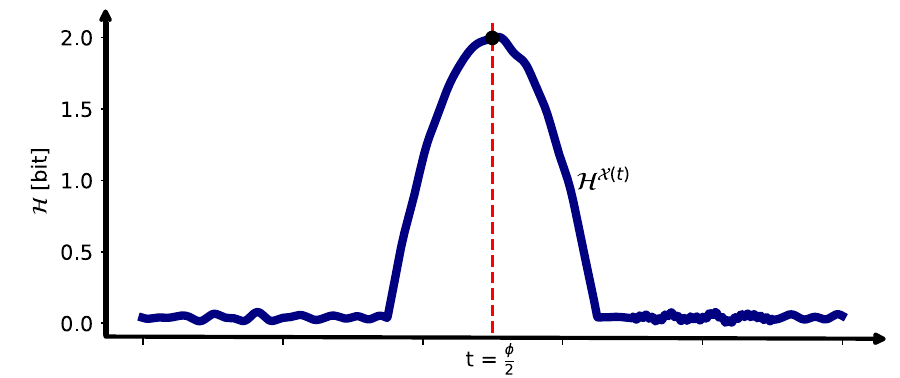}
        \subcaption{}
    \end{subfigure}
    \caption{
     The relocation of a single object being manipulated over time. (a). The trajectory of a single object is shown, with the sliding window $\phi$ applied to the signal $\mathcal{X}(t)$ representing the object's position over time. (b) The entropy $\mathcal{H}^{\mathcal{X}(t)}$ is computed by sliding the window across $\mathcal{X}(t)$ and evaluating the entropy of the distribution of positions within each windowed segment. A bell-shaped curve emerges, highlighting periods of significant positional change.
    }
    \label{fig::entropy_1}
\end{figure}

Consider the quantization of the workspace into bins. If an object is static within the window $\phi$, all position samples $x_i \in \mathbb{X}_{\phi}$ fall into a single quantization bin $k$. The probability distribution of the position variable $X$ effectively becomes a Dirac delta function $p(x) = \delta_{x,k}$, where $p(x_k)=1$ and $p(x_j)=0$ for all $j \neq k$. Substituting this into Eq. (1), the entropy becomes:
\begin{equation}
    \mathcal{H}_{static} = - \sum_{i=1}^{N_x} p(x_i) \ln p(x_i) = - (1 \cdot \ln 1) = 0
\end{equation}
Conversely, when the object exhibits motion, the trajectory traverses $M > 1$ bins. According to the principle of maximum entropy, for a support size $M$, the entropy is strictly positive, $\mathcal{H}_{dynamic} > 0$. Thus, a positive gradient $\frac{\partial \mathcal{H}^X}{\partial t} > 0$ strictly indicates a transition from a deterministic (static) state to a stochastic (moving) spatial distribution, justifying the bell-shaped profile observed in Fig. \ref{fig::entropy_1}.

Within the proposed framework, entropy analysis is primarily applied to positional signals to characterize scene dynamics effectively. Rather than computing entropy for the entire signal simultaneously, the approach involves employing a sliding temporal window $\phi$. Specifically, entropy is calculated incrementally as this window traverses the signal over time. At each incremental shift of the temporal window, we use~\eqref{eq:entropy} to generate a temporal sequence of entropy $\mathcal{H}^{\mathcal{X}(t)}$. 
% Here $t$ denotes the time step within the temporal sliding window $\phi$. 
By evaluating patterns within this entropy-derived time series, it becomes possible to discern critical fluctuations in scene dynamics, thus identifying how and when significant changes occur throughout the task execution.
% \begin{itemize}
%     % \item $\mathcal{X} := [x{\scriptstyle (t_{\phi/2}-\tfrac{\phi}{2})}, \dots, x{\scriptstyle (t_{\phi/2}+\tfrac{\phi}{2})}]$ denotes the random variable corresponding to the scene element's positional signal.
%     % \item $\mathbb{X}_{\in N_\mathrm{x}} := \{x_1, \dots, x_N\} = \{x{\scriptstyle (t_{\phi/2}-\tfrac{\phi}{2})}, \dots, x{\scriptstyle (t_{\phi/2}+\tfrac{\phi}{2})}\}$;
%     % \item $x_i$ denotes each single sample in $\mathbb{X}_{\in N_\mathrm{x}}$ where $i = 1,\dots,N$;
%     % \item $p(x_i)$ is the probability distribution for $\mathbb{S}$ derived from the empirical measurements in $\phi$;
%     % \item $\mathcal{H}^{\mathcal{X}(t_{\phi/2})}$ is the resulting entropy value.
% \end{itemize}

The probability distribution $p(x_i)$ is constructed using a histogram approach, in which each bin represents the frequency at which specific positional values occur within a designated temporal window. These positional values are systematically classified into intervals defined by a fixed quantization parameter $\zeta$, thus determining bin widths. Consequently, the height of each histogram bin corresponds directly to the number of occurrences falling within its respective quantization interval. Employing a fixed quantization interval inherently leads to a dynamic number of bins, which varies according to the range of positional data observed during motion. Evaluating the probability distribution and associated entropy of positional trajectories within the temporal window $\phi$ facilitates the identification of shifts in movement patterns through the statistical analysis of positional data. As an illustrative example, an object moving consistently at a uniform speed generates a stable entropy profile, reflecting minimal variability in positional data across the window. In contrast, when an object's movement decelerates, the positional values become increasingly predictable, resulting in a measurable decline in entropy. Such changes in entropy thus serve as indicators of dynamic variations within the scene.

A fixed quantization level, denoted by $\zeta$, enables the differentiation between spatially localized and extensive movements. For motions constrained within a limited area, variations in the window size $\phi$ will yield a probability distribution $p(x_i)$ characterized by a smaller number of bins with elevated probabilities. This outcome is a direct consequence of the object frequently occupying similar or identical positions, which results in low entropy. Conversely, for more expansive movements, the distribution $p(x_i)$ will encompass a greater number of bins, reflecting the increased diversity of the occupied positions. In the limit, this can approach a uniform distribution, corresponding to a state of high entropy. Furthermore, selecting a window size $\phi$ sufficiently large to encapsulate multiple repetitions of a given movement will cause numerous samples to possess identical or nearly identical values, thereby falling within the same quantization bin $\zeta$. Consider the example of stirring a liquid in a cup with a spoon, followed by placing the spoon on a table several centimeters away. During the cyclical stirring action, a sufficiently wide $\phi$ will capture the repetitive nature of the signal, maintaining the entropy, $\mathcal{H}(t)$, at a consistently low value. However, the subsequent action of relocating the spoon to the table introduces a transient, expansive trajectory that covers a wider range of unique spatial points. Consequently, this phase of the movement is expected to manifest as a discernible peak in the entropy profile $\mathcal{H}(t)$.

A key merit of this methodology is its capacity to detect changes in motion patterns based exclusively on the statistical distribution of positional data. This approach circumvents the reliance on velocity or acceleration profiles, which are often susceptible to inaccuracies inherent in numerical differentiation. Fig.~\ref{fig::entropy_1} can depict the one-dimensional position signal $\mathcal{X}(t)$ for an object's relocation. By computing the entropy over a sliding window $\phi$ along this signal, a temporal entropy profile $\mathcal{H}^{\mathcal{X}(t)}$ is generated, which typically exhibits a unimodal, bell-shaped characteristic. The peak of this curve, occurring at time $t_{\phi/2}$ marks the temporal center of the motion event, while its amplitude and duration provide insights into the movement's magnitude and speed. Moreover, the time derivative, $\frac{\partial \mathcal{H}^{\mathcal{X}(t)}}{\partial t}$, elucidates the dynamics of the event: a positive derivative $\frac{\partial \mathcal{H}^{\mathcal{X}(t)}}{\partial t} > 0$ indicates the initiation of motion, whereas a negative derivative  $\frac{\partial \mathcal{H}^{\mathcal{X}(t)}}{\partial t} < 0$ signals the convergence toward a new equilibrium state.

\begin{figure}[!t]
    \centering
    % \captionsetup[subfigure]{justification=centering, singlelinecheck=false, margin={1.20in, 6cm}} 
    \captionsetup{font=footnotesize}
    \begin{subfigure}[c]{0.48\textwidth}
        \centering
        \captionsetup{font=footnotesize,,margin={0.6cm,0cm}}\includegraphics[width=0.9\linewidth]{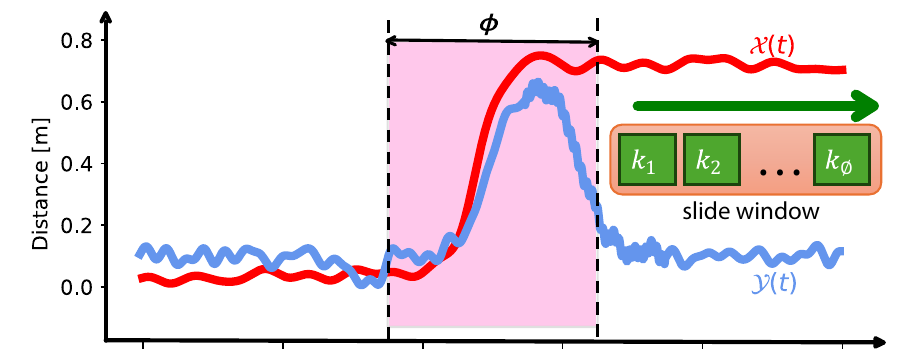}
        \subcaption{}
        \label{fig1::sub_2} 
        \includegraphics[width=0.9\linewidth]{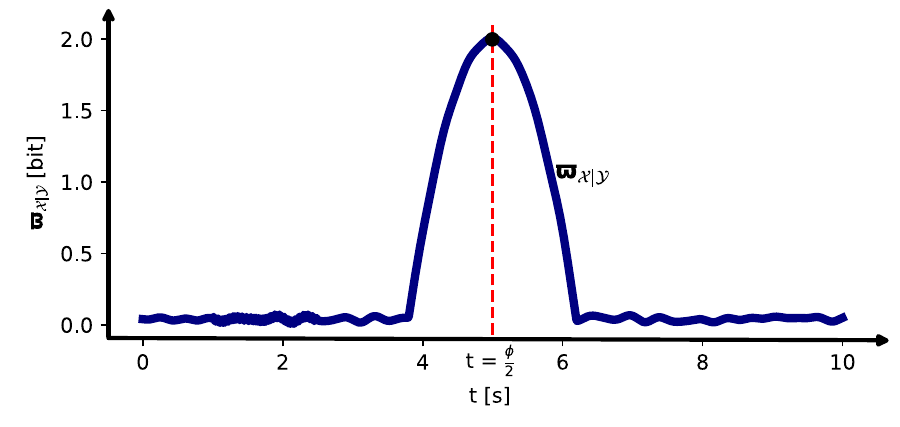}
        \subcaption{}
    \end{subfigure}
    \caption{
     The scenario of a hand moving an object over time. (a). The hand’s position is denoted by $\mathcal{Y}(t)$ and the object’s by $\mathcal{X}(t)$, with both signals captured simultaneously. (b). The mutual information $\boldsymbol{\varpi}_{\mathcal{X}|\mathcal{Y}}$ is calculated between the hand and object signals across the same windowed intervals. The resulting curve indicates time instances of coordinated motion, peaking when the hand and object move jointly.}
    \label{fig::entropy_2}
\end{figure}

A comprehensive analysis of manual tasks, however, necessitates a shift from characterizing individual scene components in isolation to examining their interactions. An interaction is fundamentally characterized by a mutual or reciprocal influence between two or more entities, implying a shared exchange of information. Shannon's information theory again provides a powerful framework for this purpose, enabling the quantification of information shared between multiple signals. To measure this shared information, we employ the concept of mutual information ($\boldsymbol{\varpi}$), a metric that quantifies the statistical dependency between two random variables. Conceptually, if the entropies of two distinct random processes, $\mathcal{X}$ and $\mathcal{Y}$, are represented by the sets $\mathcal{H}^{\mathcal{X}}$ and $\mathcal{H}^{\mathcal{Y}}$, the $\boldsymbol{\varpi}$ corresponds to the measure of their intersection. It is calculated as follows
\begin{equation}
\boldsymbol{\varpi}_{\mathcal{X}|\mathcal{Y}} = \mathcal{H}^{\mathcal{X}} + \mathcal{H}^{\mathcal{Y}} - \mathcal{H}^{\mathcal{X},\mathcal{Y}}, 
\label{eq:mi}
\end{equation}
where $\boldsymbol{\varpi}_{\mathcal{X}|\mathcal{Y}}$ is called \emph{joint entropy} and represents the union of the two sets:
\begin{equation}
\label{eq:joint_e}
\boldsymbol{\varpi}_{\mathcal{X}|\mathcal{Y}} = - \sum_{i=1}^{N_\mathrm{x}} \sum_{j=1}^{N_\mathrm{y}} p(x_i,y_j) \cdot \ln{p(x_i, y_j)},
\end{equation}
where $x_i$ and $y_j$ are discrete measurements from $\mathbb{X}_{\phi \in N_\mathrm{x}}$ and $\mathbb{Y}_{\phi \in N_\mathrm{y}}$, respectively, and $p(x_i,y_j)$ is the joint probability of both events occurring simultaneously in the time window $\phi$. Based on \eqref{eq:mi}, we can conclude that if $\boldsymbol{\varpi}_{\mathcal{X}|\mathcal{Y}} = 0$, $\mathcal{X}$ is independent of $\mathcal{Y}$ and vice versa, due to the symmetry of this measure. In other words, knowledge about the value of one variable provides no information about the other variable. Since we are interested in observing the dynamics of $\boldsymbol{\varpi}$ over time, we apply the same approach described earlier, shifting the time window $\phi$ along the considered signals to obtain $\boldsymbol{\varpi}_{\mathcal{X}(t)|\mathcal{Y}(t)}$.

The application of $\boldsymbol{\varpi}$ to the position signals of two objects provides a quantitative measure of their kinematic dependency. Within the context of manual activities, this allows for a robust determination of whether two elements, such as a hand and a target object, are coupled in their motion. A significant advantage of $\boldsymbol{\varpi}$ is its ability to capture both linear and non-linear statistical relationships, offering a more comprehensive analysis of complex dependencies than methods relying solely on velocity-profile correlations. For instance, in a scenario where a hand transports multiple items on a tray, $\boldsymbol{\varpi}$ can correctly identify the hand and objects as a single kinematic unit, even in the presence of minor, non-linear sliding or oscillatory motions of the items during transport. Fig.~\ref{fig::entropy_2} illustrates a canonical grasp-and-move task, involving the one-dimensional position signals of a hand, $\mathcal{Y}(t)$, and an object, $\mathcal{X}(t)$. Computing the time-varying $\boldsymbol{\varpi}_{\mathcal{X}|\mathcal{Y}}$ using a sliding window reveals a distinct unimodal profile that corresponds precisely to the interval of coordinated movement. Note that $\boldsymbol{\varpi}$ remains at or near zero during the pre-grasp approach and post-release retreat phases. This property is critical for disambiguating true interaction from incidental proximity in cluttered environments, such as when a hand moves near one object to manipulate another. To accommodate analysis in three-dimensional space, our framework aggregates the information across all spatial axes. The total mutual information between a hand $h \in \mathbb{H}_{\in N_h}$ and an object $o_i \in \mathbb{O}_{\in N_o}$, can be denoted as $\boldsymbol{\varpi}_{h,o_i}$, is therefore calculated as the sum of the $\boldsymbol{\varpi}$ values computed independently for each of the three orthogonal directions.

Furthermore, we formulate the detection of physical grasping as a maximization problem of information sharing between the hand and the object. The mutual information $\varpi_{X|Y}$ between a hand trajectory $Y$ and an object trajectory $X$ is maximized when a rigid kinematic constraint (e.g., a stable grasp) is established.

The mutual information can be expressed as the reduction in the uncertainty of $X$ given knowledge of $Y$:
\begin{equation}
    \varpi_{X|Y} = \mathcal{H}^X - \mathcal{H}^{X|Y}
\end{equation}
In a \textit{Coupled-Motion} scenario (i.e., a stable grasp), the object's position $X$ undergoes a rigid body transformation determined by the hand's position $Y$. This can be modeled as $X = R \cdot Y + T + \epsilon$, where $R$ and $T$ are rotation and translation matrices, and $\epsilon$ represents negligible sensor noise. Under this deterministic mapping, the conditional entropy $\mathcal{H}^{X|Y}$ approaches zero (or the entropy of the noise floor). Consequently, the mutual information approaches the self-entropy of the object's path:
\begin{equation}
    \lim_{\text{coupling} \to \text{rigid}} \varpi_{X|Y} \approx \mathcal{H}^X
\end{equation}
In contrast, if $X$ and $Y$ are independent (no interaction), their joint probability factorizes as $p(x,y) = p(x)p(y)$, leading to $\varpi_{X|Y} = 0$. Therefore, a high value of $\varpi_{X|Y}$ serves as a robust theoretical indicator of kinematic coupling.

\begin{figure*}
    \centering
    \includegraphics[width=0.9\linewidth]{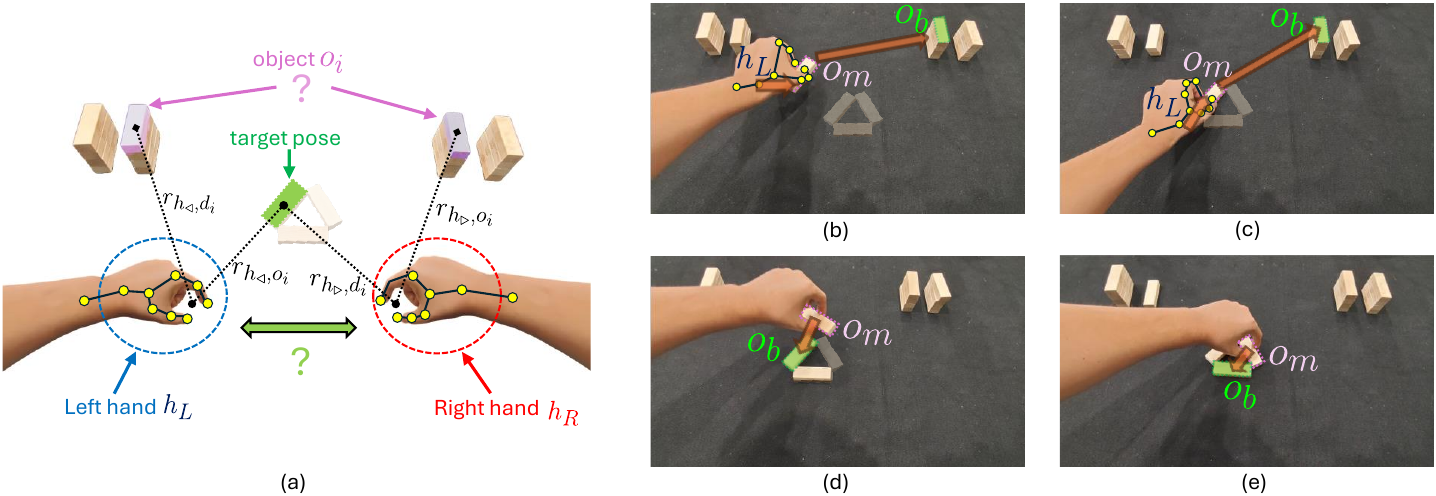}
    \caption{(a) The conceptual representation of the dual-hand selection policy. The framework depends on the priority of the Left hand $h_L$ or the Right hand $h_R$, which is optimal for interacting with the manipulated object $o_m$ to move it to the target pose. (b) denotes \texttt{Coupled-Motion} integration between only the left-hand $h_L$ and one manipulated Jenga block $o_m$. (c) denotes the \texttt{Docked} interaction between $h_L$ and one Jenga block $o_m$. (d) denotes \texttt{E-OO} interaction between the manipulated jenga block $o_{m}$ and one background jenga block $o_b$ on the table. (e) denotes the \texttt{T-OO} interaction between one manipulated Jenga block $o_{m}$ and the current background Jenga block $o_b$ when the hand is shaken to make the building blocks shift slightly near the target position.}
    \label{fig:topo_4_relationship}
\end{figure*}

\subsection{Scene Graph Generation}
We employ a scene graph, which is a directed graph data structure, to represent the system state and the relationships between its components. Formally, a graph can be denoted as $\texttt{SR}:= (\mathcal{G}, \mathcal{F})$ where $\mathcal{G}:=(\mathbb{V}_{\in N_\mathrm{v}}, \mathbb{R}_{\in N_\mathrm{r}}, \mathbb{E}_{\in N_\mathrm{e}})$ is a tuple in which $\mathbb{V}_{\in N_\mathrm{v}}$ is a set with $N_v$ number nodes, $\mathbb{R}$ is the set of relationships between the nodes with number $N_\mathrm{r}$, and $\mathbb{E}$ is a set of edges with number $N_\mathrm{e}$, and $\mathcal{F}:= f \sim \mathbb{K}_{\phi \in N_x}$ is a combined set of frames which we extract from the learning video during a temperoal sliding window. In our framework, the nodes $\nu_i:=(\lambda_i, \Lambda_i) \in \mathbb{V}_{\in N_\mathrm{v}}$ in which  $\Lambda_i$ is a set of attributes and $\lambda_i$ is its class identity, correspond to the entities within the scene, specifically hands and objects.  For our application, the sole attribute is the entity's 6D pose $\mathbf{q}_i \in \reals^6$. The directed edges $e_i \in \mathbb{E}_{\in N_\mathrm{e}}$ denote the interactions between these entities. Within the context of uni-manual tasks, interactions are classified into two principal categories:
\begin{itemize}
    \item \textit{\textbf{H}and-\textbf{O}bject (HO) Interactions}: denotes the relationship that pertains to the coupling between a hand and a manipulated object.
    \item \textit{\textbf{O}bject-\textbf{O}bject (OO) Interactions}: denotes the relationship between one moving manipulated object in hand and another static object in the background.
\end{itemize}

The detection of an interaction between any two entities results in the instantiation of a corresponding directed edge connecting their representative nodes in the scene graph. The scene graph generation procedure is initiated by identifying \textit{HO} interactions. This sequence is predicated on the hand's role as the primary effector capable of inducing state changes within the environment. Subsequently, the algorithm evaluates potential \textit{OO} interactions. The secondary analysis is strictly conditional that it is executed only if a \textit{HO} interaction has been successfully identified in the initial phase. In the absence of such a detection, the \textit{OO} interaction analysis stage is bypassed.

In the following, we will introduce the scene graph generation procedure for a single hand, $h_k \in \mathbb{H}_{\in N_{\mathrm{h}}}$ at a given frame $k$. The adoption of an information-theoretic framework substantially simplifies the detection and identification of interactions, yielding an intuitive and computationally straightforward process. The constituent steps of this procedure are elaborated upon in the subsequent subsections.

\subsubsection{Hand–Object Interactions Detection} We define two classes of Hand-Object ($\textit{HO}$) interactions as following:
\begin{itemize}
    \item \texttt{Coupled-Motion}: denotes interaction by the active displacement of an object by the hand, indicating that the hand holds the object and moves simultaneously.
    \item \texttt{Docked}: denotes a state of relative static contact between the hand and object, wherein the hand is in physical contact with an object but no joint movement is observed.
\end{itemize}

If the proximity condition is satisfied, the interaction is classified based on the mutual information, $\boldsymbol{\varpi}_{h,o_i}$. An active \texttt{Coupled-Motion} is declared if $\boldsymbol{\varpi}_{ho_i}$ is greater than a near-zero threshold $\alpha_{\boldsymbol{\varpi}}$, signifying strong kinematic coupling. Conversely, a \texttt{Docked} interaction is established or maintained if either the mutual information is concurrently decreasing and below the threshold (i.e., $\frac{\partial \boldsymbol{\varpi}_{h,o_i}}{\partial t} < 0$ and $\boldsymbol{\varpi}_{h,o_i} < \gamma{\boldsymbol{\varpi}}$), or if a \texttt{Docked} state was already present in the previous frame $k-1$. Any established interaction is terminated once the distance $\bar{r}_{h,o_i}$ subsequently exceeds the threshold $r^{\mathrm{th}}_{h,o} \in \reals_{\ge 0}$.

A key aspect of our formulation is that a \texttt{Docked} state can only be entered subsequent to a \texttt{Coupled-Motion} phase. This constraint is by design and prevents the erroneous detection of incidental contact as a salient interaction before purposeful manipulation has occurred. This approach significantly enhances the robustness of the system, particularly in cluttered environments where transient, non-manipulative contact is common. The detection procedure is executed iteratively for each object $o_k \in \mathbb{O}_{\in N_o}$ present in the scene, ordered by increasing proximity to the hand $h_k$. The search for hand-object interactions terminates immediately upon the first positive detection, after which the algorithm proceeds to the object-object interaction analysis stage.

\subsubsection{Object–Object Interactions Detection}
We define two classes of Object-Object ($OO$) interactions in static as following:
\begin{itemize}
    \item \texttt{Efficient OO (E-OO)}: denotes an interaction between a manipulated object $o_m$ and a stationary background object $o_b$ that is identified as being stable, intentional, and integral to the execution of a task. This type of interaction is considered essential and is subsequently mapped into robot command primitives during plan generation.
    \item \texttt{Transitory OO (T-OO)}: denotes incidental interaction between a manipulated object $o_m$ and a stationary background object $o_c$ that is identified as being unstable, transitory, and not representative of the core activity. These interactions are considered non-essential and are filtered out during the task segmentation process, meaning they are not translated into robot commands.
\end{itemize}

A prerequisite for any static $OO$ interaction is spatial proximity. Specifically, the average distance $\bar{r}_{o_m,o_b}$ between the involved objects must be less than a predefined threshold $r^{\text{th}}_{\textit{o,o}}$ as $\bar{r}_{o_m,o_b} < r^{\text{th}}_{\textit{oo}}$. If this condition is met, a \texttt{E-OO} interaction is established or confirmed if either of two conditions holds: (i) a \texttt{Docked} interaction exists between the hand $h$ and the manipulated object $o_m$, or (ii) an \texttt{E-OO} interaction between $o_m$ and $o_b$ was already present in the preceding frame, $k-1$. This formulation ensures that if, following a manipulation, the hand holds $o_m$ stationary in the vicinity of $o_b$, the resulting static proximity is correctly classified as a significant interaction.

In scenarios where the hand is actively manipulating $o_m$ (i.e., a \texttt{Coupled-Motion} interaction is present), a different approach is needed to evaluate interactions with stationary objects, as mutual information is ill-suited for cases where one object's position entropy is zero. We therefore exploit the entropy of the average distance between the objects, $\mathcal{H}(\bar{r}_{o_m, o_b})$, as this metric effectively captures the stability of their spatial relationship. The classification of the interaction hinges on the time derivative of this distance entropy. Specifically, the $OO$ interaction is deemed \textit{efficient} if the entropy is decreasing ($\frac{\partial \mathcal{H}(\bar{r}_{o_m,o_b})}{\partial t} < 0$); otherwise, it is classified as \textit{Transitory}.

The rationale for this criterion is that a decreasing distance entropy indicates a transition towards a stable configuration, where the distance between the objects becomes more predictable and thus signifies an intentional interaction. Conversely, an increasing entropy profile suggests a transient encounter. Once an interaction is classified as \textit{Expective}, it retains this label to reflect the establishment of a new equilibrium state, even if $\mathcal{H}(\bar{r}_{o_m,o_b})$ subsequently fluctuates. This \textit{Expective} classification is revoked only if the objects' separation  $\bar{r}_{o_m,o_b}$  exceeds the proximity threshold $r^{\text{th}}_{\textit{o,o}}$. For instance, if an object $o_{i}$ is merely moved past another object $o_{i-1}$, their distance entropy will increase after their initial proximity, correctly identifying the interaction as \textit{transitory} before it is terminated by exceeding the distance threshold.

This evaluation is performed iteratively for the specific background object $o_b$ in the scene, ordered by increasing distance from $o_m$, and the process terminates upon the first detection of any $OO$ interaction. By requiring all static $OO$ interactions to involve the manipulated object, our framework avoids the generation of spurious edges between incidental on-table objects. This ensures that the resulting graph topology remains consistent and robust, accurately representing the core task regardless of environmental clutter.

After the whole process of the integration section is finished, we can assume $\mathcal{SR}[k]$ to be on the topology among those illustrated in Fig.~\ref{fig:topo_4_relationship} from (b) to (e). In the case where both a hand-object ($HO$) and a subsequent \texttt{RS-OO} interaction are detected, the graph at instant $t$ is fully specified as $\mathbb{V}_t := \{h, o_1, o_2\}$ in which $h = (\mathcal{L}_h, \mathbf{p}_h{\scriptstyle (t)})$, $o_1 = (\mathcal{L}_{o_1}, \mathbf{p}_{o_1}(t))$, and $o_2 = (\mathcal{L}_{o_2}, \mathbf{p}_{o_2}{\scriptstyle (t)})$. Note that here we use $\mathcal{L}$ to classify different objects in the label.  $\mathbb{E}_t = \{e_{h \rightarrow o_1}, e_{o_1 \rightarrow o_2}\}$ denotes the directed edges of the interaction chain. $\mathbb{R}_t = \{r_{h \rightarrow o_1}, r_{o_1 \rightarrow o_2}\}$, where the $HO$ relationship $r_{h \rightarrow o_1}$ is annotated with its type and the corresponding mutual information $\varpi_{h,o_1}(t)$, and the $OO$ relationship $r_{o_1 \rightarrow o_2}$ is labeled with its static interaction type. In the simpler scenario where no static $OO$ interaction is identified, the $\mathbb{V}, \mathbb{R}, \mathbb{E}$ composed in graph $\mathcal{G}[k]$ can be reduced to $\mathbb{V} = \{h, o_1\}$, $\mathbb{E} = \{e_{h \rightarrow o_1}\}$, and $\mathbb{R} = \{r_{h \rightarrow o_1}\}$, capturing only the direct hand-object relationship.

\subsubsection{Dynamic Learning Policy of Dual-Hand Selection}
% To facilitate the dual-hand in robot manipulation for computational allocation of resources and operation period to improve the manipulation efficiency compared to the same manipulation task deployed via a single robot arm, we come up with a dynamic policy $\pi$ for learning that \textcolor{red}{TBD}.
In dual-arm manipulation tasks, the efficient allocation of actions to each end-effector is paramount for optimizing execution time, minimizing energy consumption, and avoiding inter-arm collisions. To address this challenge, we introduce a dynamic learning policy for dual-hand selection that determines the optimal hand to execute a grasping action based on the spatial configuration of the workspace. This policy is not only grounded in principles of kinematic efficiency but is also refined through observational learning from human demonstrations, ensuring that the robot's behavior is both performant and intuitively aligned with human strategies.

The core of our policy is formulated around a concise state representation derived from the geometric relationships between the hands, the object sources, and the target placement pose. For a given task requiring the placement of an object $O_i$, we define the state $s_t$ at time $t$ by four key Euclidean distances, as illustrated in Figure 1(a). Here, $r_{h_{\triangleleft}, d_i}$ denotes the distance between left-hand $h_L$ to one jenga block on the left-hand side jenga block source; $r_{h_{\triangleleft}, o_i}$ denotes the distance between the left hand $h_L$ to the target pose in the middle of two hands; $r_{h_{\triangleright}, d_i}$ denotes the distance between right-hand $h_R$ to one jenga block on the right-hand side jenga block source; $r_{h_{\triangleright}, o_i}$ denotes the distance between the right hand $h_R$ to the target pose in the middle of two hands. The decision-making process of the policy $\pi\left(s_t\right)$ designed to select an action $a_t \in \{ \text{use\_left\_hand}, \text{use\_right\_hand} \}$ that promotes an efficient, crossing-arm strategy. This strategy is predicated on the observation that it is often more efficient for a hand to retrieve an object from the contralateral (opposite) side relative to the final placement location. This minimizes the total path length for the combined action of reaching, grasping, and placing.

The policy is defined by the following rule: If the target placement pose is closer to one side of the workspace, the hand on the opposite side is selected to perform the grasp. For instance, if the target pose is closer to the right hand's side of the workspace, the left hand is chosen to retrieve an object from the left-side source. This logic is formalized as:

\begin{equation}
a_t = 
\begin{cases} 
\text{use\_left\_hand} & \text{if } r_{h_{\triangleright}, o_i} < r_{h_{\triangleleft}, o_i} \\
\text{use\_right\_hand} & \text{if } r_{h_{\triangleleft}, o_i} \le r_{h_{\triangleright}, o_i}
\end{cases}
\end{equation}

% \[
% \begin{array}{|c|c|}
% \hline
% \textbf{Condition} & \textbf{Action} \\
% \hline
% r_{h_{\triangleright}, o_i} < r_{h_{\triangleleft}, o_i} & \text{use\_left\_hand} \\
% r_{h_{\triangleleft}, o_i} \leq r_{h_{\triangleright}, o_i} & \text{use\_right\_hand} \\
% \hline
% \end{array}
% \]

To ensure the policy aligns with effective and natural human movements, we introduce a learning mechanism that provides feedback based on observations from human demonstration videos. For each decision point in a task, the action taken by the human demonstrator, $a_t^*$ is extracted from the video data and serves as the ground truth. The action $a_t$ selected by our policy is then compared against this ground truth.

We define a reward function $R\left(a_t, a_t^*\right)$ to quantify the concordance between the policy's action and the human's action. This function provides a positive reward (a bonus) when the policy's choice matches the human's, and a negative reward (a penalty) when it deviates. This can be expressed as:
\begin{equation}
R\left(a_t, a_t^*\right)= \begin{cases}+R_{\text {bonus }}, & \text { if } a_t=a_t^* \\ -R_{\text {penalty }}, & \text { if } a_t \neq a_t^*\end{cases}
\end{equation}
where $R_{\text {bonus }}>0$ and $R_{\text {penalty }}>0$ are scalar values. This reward signal is integrated into the broader learning framework of GF-VLA as shown in Fig.~\ref{fig:VLA_overview}. It acts as a strong inductive bias, guiding the agent to learn not just what actions to perform, but how to perform them in a bimanually efficient manner. Over time, this feedback mechanism allows the system to refine its decision-making, reinforcing strategies that are both computationally optimal and consistent with observed human intelligence, thereby improving the overall efficiency and naturalness of the dual-arm manipulation.

\section{Graph-Fused VLA (GF-VLA)}

\begin{figure*}[!t]
    \centering
    \includegraphics[width=0.95\linewidth]{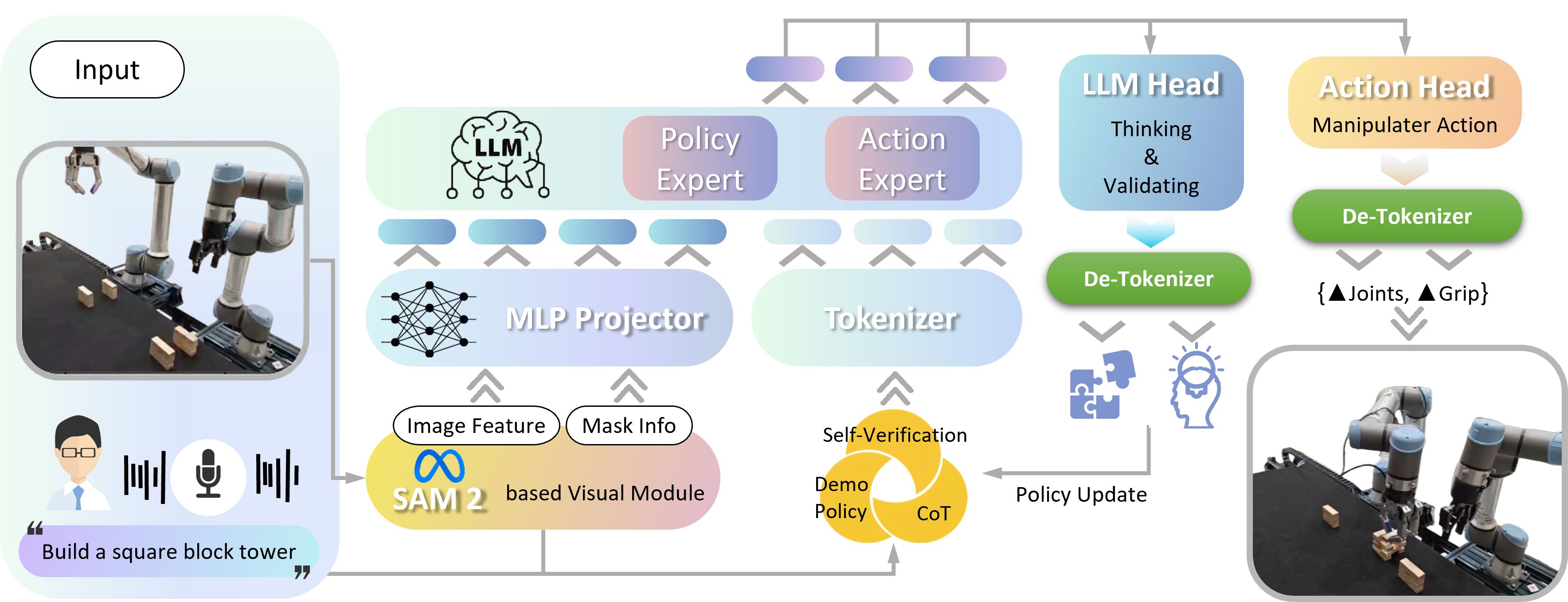}
    \caption{Policy transfer from a single human demonstration to a novel dual-arm robotic assembly task. The framework processes multimodal inputs, including language commands and visual scene data, using a SAM 2-based module to extract and project features into a shared embedding space. At its core, a unified Large Language Model (LLM) employs a dual-head structure: the LLM Head performs high-level semantic planning and validation using Chain-of-Thought (CoT) and self-verification, while the Action Head generates low-level, executable actions for manipulators. This integrated design enables the robot to de-tokenize abstract reasoning into physically grounded joint and gripper commands, translating high-level goals into robust, real-world execution.}
    \label{fig:VLA_overview}
\end{figure*}

In this section, we introduce the \textit{GF-VLA} framework as shown in Figure \ref{fig:VLA_overview}, which combines information-theoretic abstraction and vision-language-action reasoning to enable generalizable and interpretable robotic manipulation. We first describe the overall system architecture composed of two stages: policy generation (Demo policy) from human demonstration and execution via CoT-guided policy reasoning. We then detail how information theory is employed to extract structured behavior graphs from raw demonstration videos, providing symbolic priors that encode key semantic and temporal task elements. Next, we present our unified dual-head VLA model, which fuses multimodal inputs—RGB-D perception and semantic instructions—within a shared transformer backbone to jointly generate executable actions and structured policies. To support interpretable planning and robust execution, we incorporate a chain-of-thought (CoT) reasoning mechanism with embedded self-verification, enabling explicit articulation of subgoals and dynamic policy adjustments. Finally, we describe a parameter-efficient multi-head fine-tuning strategy based on Low-Rank Adaptation (LoRA), which independently optimizes the LLM head and action head for their respective roles under a unified semantic representation. 

%% Use \subsection commands to start a subsection.
\subsection{System architecture}
\label{framework}

We present a detailed overview of our system framework in Figure~\ref{fig:overview_gf_vla}, which consists of two major stages: (1) Policy Generation from Human Demonstration, and (2) VLA-based Policy Execution with Chain-of-Thought Reasoning. Together, these stages enable robots to understand, learn, and plan manipulation tasks in unstructured environments through semantic scene abstraction and reasoning-augmented policy design. We emphasize that this stage constitutes a policy transfer mechanism. The system leverages the pre-trained VLA backbone to adapt the demonstrated strategy to the robot's embodiment, distinct from training new behavior policies from a single example.

In the first stage, raw human demonstration videos are processed to extract structured information-theoretic behavior graphs and temporal keyframes. These representations encode spatial and semantic task elements across time. A dedicated Policy Agent interprets and reasons over the graph-frame structures to generate a robot-understandable task policy, consisting of a sequence of subtasks and an associated behavior reasoning trace, using CoT and Self-verification framework.

The second stage takes as input the CoT policy generated from human demonstration, along with live RGB visual input and spoken instructions from the user. Our GF-VLA model fuses these multimodal inputs to generate an updated robot manipulation policy that reflects both the original human intent and the current scene context. The model produces two outputs: an action head, responsible for generating low-level executable robot commands; and an LLM head, which performs policy: hierarchical task planning, subtask decomposition, and CoT-based reasoning. During execution, the system continuously performs self-verification, allowing the robot to assess task progress, detect inconsistencies, and locally replan when deviations occur—ensuring robust, interpretable, and adaptive behavior across complex multi-step manipulation tasks.

%% Use \subsubsection, \paragraph, \subparagraph commands to 
%% start 3rd, 4th and 5th level sections.
%% Refer following link for more details.
%% https://en.wikibooks.org/wiki/LaTeX/Document_Structure#Sectioning_commands

\subsection{Unified Dual-Head Architecture for Vision-Language-Action Reasoning}
\label{sec:model_architecture}

% Our system architecture builds upon the standard VLA framework, incorporating four core components: (1) a visual module that segments the RGB image and encodes the RGB-D image with mask inputs into patch-level embeddings;
% (2) a lightweight MLP projector for cross-modal token alignment;
% (3) a language module built upon a pretrained 7B-parameter LLaMA 2 model by Open X-Embodiment dataset \cite{o2024open, kim2024openvla}, augmented with a reasoning mechanism for hierarchical policy generation, chain-of-thought (CoT) decomposition, and self-verification;
% (4) two task-specific output heads: an LLM Head for structured semantic planning and interpretable reasoning, and an Action Head for low-level motion control. Unlike the original VLA design, GF-VLA integrates both heads into a unified transformer encoder, enabling parallel reasoning and action generation under a shared semantic representation. For speech-based prompting, we further include a deep learning-based Gladia Automatic Speech Recognition (ASR) module to transcribe user voice commands into text.

Building upon the foundational VLA paradigm \cite{o2024open, kim2024openvla} that aligns visual patch embeddings with a pre-trained LLaMA 2 backbone, our system introduces a novel unified dual-head transformer architecture. Deviating from conventional designs that separate reasoning and control, GF-VLA integrates both an LLM Head and an Action Head within a shared transformer encoder. This unified structure facilitates parallel processing, enabling the model to perform high-level semantic planning—including Chain-of-Thought (CoT) decomposition and self-verification simultaneously with low-level motion control. To further enhance usability, the framework is augmented with a Gladia-based ASR module, translating vocal commands directly into the text-conditioned workflow.

The architecture of GF-VLA is illustrated in Figure\ref{framework}. Visual inputs are obtained via a top-down RGB-D camera, while verbal instructions are transcribed and transformed into prompts for the language module. Visual patch embeddings and binary masks are projected through a two-layer MLP before being fused with the prompt tokens. The language module is built upon a pretrained 7B-parameter LLaMA 2 backbone and is augmented with reasoning capabilities that support hierarchical policy generation, Chain-of-Thought (CoT) decomposition, and self-verification for error-aware planning. During task execution, the model generates a one-time structured reasoning trace via the LLM Head and produces a continuous stream of control actions at 5Hz via the Action Head. The action stream encodes both Cartesian poses and gripper states, which are decoded and dispatched to the corresponding robotic arms.

\begin{figure*}
    \centering
    \includegraphics[width=0.9\linewidth]{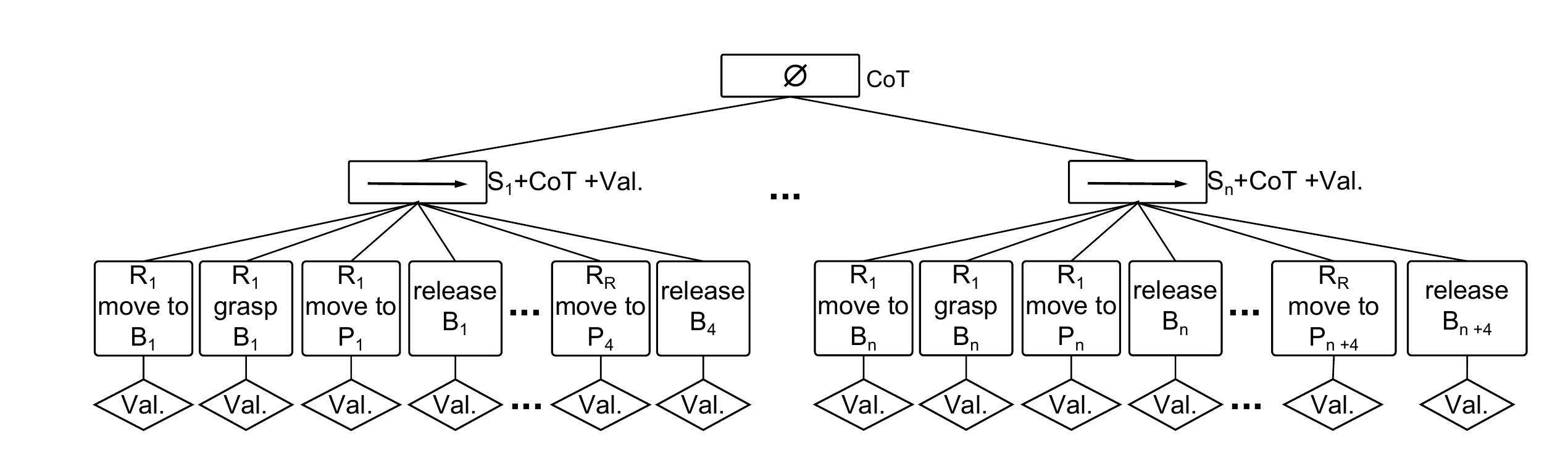}
    \caption{Behavior Tree with CoT reasoning and self-verification. The root node ($\emptyset$) generates a CoT-guided policy, decomposed into $n$ subtasks. Each subtask includes validation for action correctness before proceeding.}
    \label{fig:BT}
\end{figure*}

\subsection{Chain-of-Thought Guided Semantic Policy Planning}
\label{sec:policy_with_cot}

\begin{tcolorbox}[promptstyle, fontupper=\ttfamily\small, title=Task planning with CoT and self-verification:, label=box:cot_plan]
\textcolor{red}{"Task Planning"}:\\
\{
  \textcolor{red}{"node"}: "PickObjDual",\\
  \textcolor{red}{"param"}: "left: block\_A, right: block\_B",\\
  \textcolor{red}{"reason"}: "Grasp base blocks for letter R.",\\
  \textcolor{red}{"verify"}: "Grippers closed; correct blocks confirmed."
\},\\

\textcolor{gray}{...}\\
\\
\{
  \textcolor{red}{"node"}: "PlaceObjDual",\\
  \textcolor{red}{"param"}: "left: block\_E, right: block\_D",\\
  \textcolor{red}{"reason"}: "Form diagonal leg of R.",\\
  \textcolor{red}{"verify"}: "Pose verified; structure stable."
\},\\
\textcolor{red}{"Final Analysis"}: "R assembled correctly; no regrasp needed."
\end{tcolorbox}

Robots operating in real-world environments must produce behavior plans that exhibit logical structure, causal coherence, and alignment with human commonsense reasoning. However, traditional planning modules based on LLMs often adopt a "fast thinking" paradigm—favoring single-step, reactive outputs over deliberate, structured plans. While suitable for simple or atomic actions, such approaches fall short when handling multi-stage manipulation tasks that require temporally extended reasoning and subgoal decomposition.

To address this, we incorporate Chain-of-Thought (CoT) reasoning into the GF-VLA’s policy generation process \cite{wei2022chain}. The CoT framework encourages the Planning Agent to explicitly articulate the reasoning process, slowing down decision-making and breaking down high-level goals into interpretable subgoals. This not only enhances the logical consistency and correctness of the generated plans, but also improves their interpretability and auditability. Each intermediate reasoning step is articulated in natural language, facilitating downstream plan verification, debugging, and user understanding.

As illustrated in the box below, the CoT-enhanced Planning Agent produces a structured behavior tree aligned with human task logic, where each node contains not only the action type and parameters but also step-wise reasoning and self-verification criteria. 
In the example of constructing the letter “R”, the robot first performs a dual-arm grasp action (\texttt{PickObjDual}) to simultaneously pick up two foundational blocks, guided by the structural intent to build the vertical backbone. It then executes a coordinated placement (\texttt{PlaceObjDual}) to complete the diagonal leg. Each action node includes a reasoning statement describing the subgoal and a verification clause to confirm execution success based on sensor feedback (e.g., gripper status, object pose, and scene stability). 
This CoT-guided, interpretable plan supports temporal consistency, symbolic understanding, and robust execution in complex bimanual manipulation tasks.

\subsection{Parameter-Efficient Multi-Head Fine-Tuning with LoRA}
\label{sec:training}

To enable efficient learning and head specialization, we adopt Low-Rank Adaptation (LoRA) \cite{hu2021lora} for parameter-efficient fine-tuning. Separate LoRA adapters are inserted for the LLM Head and Action Head within the shared LLaMA 2 transformer blocks. This setup allows us to train both heads independently with distinct data modalities and task-specific objectives while maintaining a unified multimodal encoder. During training, the visual encoder and projector are frozen, and only the LLaMA backbone and LoRA adapters are updated.

The LLM Head is trained using policy-level supervision derived from human demonstrations. Following Merlo et al.~\cite{merlo2025exploiting}, we collected a total of 250 RGB videos from ten participants performing two types of block manipulation tasks: letter-based symbolic assembly and tower construction. Each video was segmented into graph-based keyframes and paired with expert-annotated task strategies. Among them, 125 demonstrations were used to train the LLM Head by supervising next-token prediction for plan steps and Chain-of-Thought (CoT) reasoning. The remaining 125 demonstrations were reserved for downstream evaluation of policy generation in Subsection 5.2.

The Action Head is fine-tuned using a subset of a 240-trial dataset collected from a dual-arm robotic platform consisting of a UR5e (with a Robotiq 2F-85 parallel gripper) and a UR10e (with a Barrett Hand BH282 dexterous gripper), both mounted on a shared base as shown in Figure~\ref{fig::exp_setup}. A top-mounted Intel RealSense D435i RGB-D camera provides a bird’s-eye view of the workspace for real-time perception and object tracking.
All trials were conducted in a bimanual setup, with both arms executing coordinated motions. 
Four task categories were designed to capture core manipulation capabilities:

    \paragraph{Shape Generalization:} The robot was instructed to pick up a specific block shape—cuboid, triangular prism, or cube—and place it at a predefined location. This assessed the model’s ability to generalize grasp strategies across geometries. Both arms participated in precise handoff and pose alignment.
    
    \paragraph{Spatial Relation via Ambiguous Instruction:} Under-specified commands (e.g., “Place the long block next to the square block”) were given, requiring the system to interpret semantic spatial relations. One arm placed the reference object, while the other executed a relational placement, testing the system's semantic reasoning.
    
    \paragraph{Absolute 6D Pose Execution:} The model was given explicit Cartesian and rotational goals for block placement. Both arms collaborated to meet the precision requirements of the 6D target pose, particularly for orientation-sensitive placements.
    
    \paragraph{Relative Pose Execution:} Commands specified relational configurations (e.g., “Place the triangle block intersecting the cube from above”), which the system translated into relative 6D goals. Each arm executed its role to realize the composite spatial layout.

Each task type was evaluated using three different block shapes, with 20 repetitions per shape. This yielded a total of $4$ tasks $\times$ $3$ shapes $\times$ $20$ trials = 240 trials. All trajectories were recorded as synchronized dual-arm motion sequences with role-specific task assignments and corresponding 6-DoF end-effector targets. Supervision was applied using regression losses on Cartesian poses and gripper states.
Among the 240 trials, 120 trials were used to fine-tune the Action Head, while the remaining 120 trials were held out and used for evaluating task execution performance, as reported in Subsection~\ref{sec:task_execution_assessment}.

We train the two heads independently by alternating between policy and action batches, activating the corresponding LoRA branch and loss function. Using this approach, GF-VLA was fine-tuned within 40 hours on a single NVIDIA RTX 4090 GPU. During deployment, the model achieves a 7.56-second average instruction inference time under bfloat16 precision. This architecture enables interpretable reasoning and robust dual-arm manipulation under diverse spatial configurations.

\section{Experiments}

\begin{figure}[!t]
    \centering
    \includegraphics[width=0.98\linewidth]{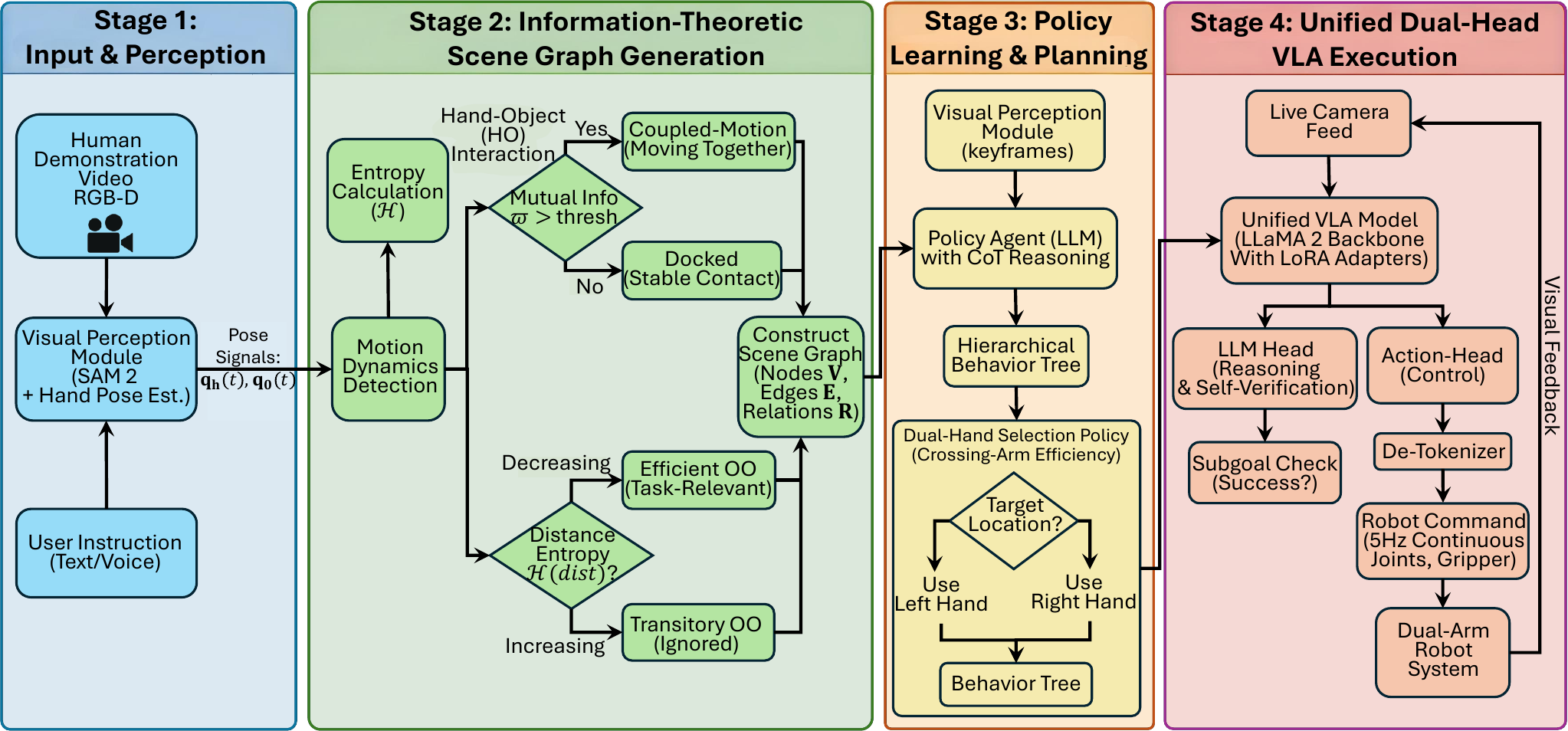}
    \caption{\textcolor{blue}{Overview of the Graph-Fused Vision-Language-Action (GF-VLA) framework.}}
    \label{fig::workflow_all}
\end{figure}

\begin{figure}[!t]
    \centering
    % \captionsetup[subfigure]{justification=centering, singlelinecheck=false, margin={1.20in, 6cm}} 
    \captionsetup{font=footnotesize}
    \begin{subfigure}[c]{0.48\textwidth}
        \centering
        \captionsetup{font=footnotesize,,margin={0cm,0cm}}\includegraphics[width=0.8\linewidth]{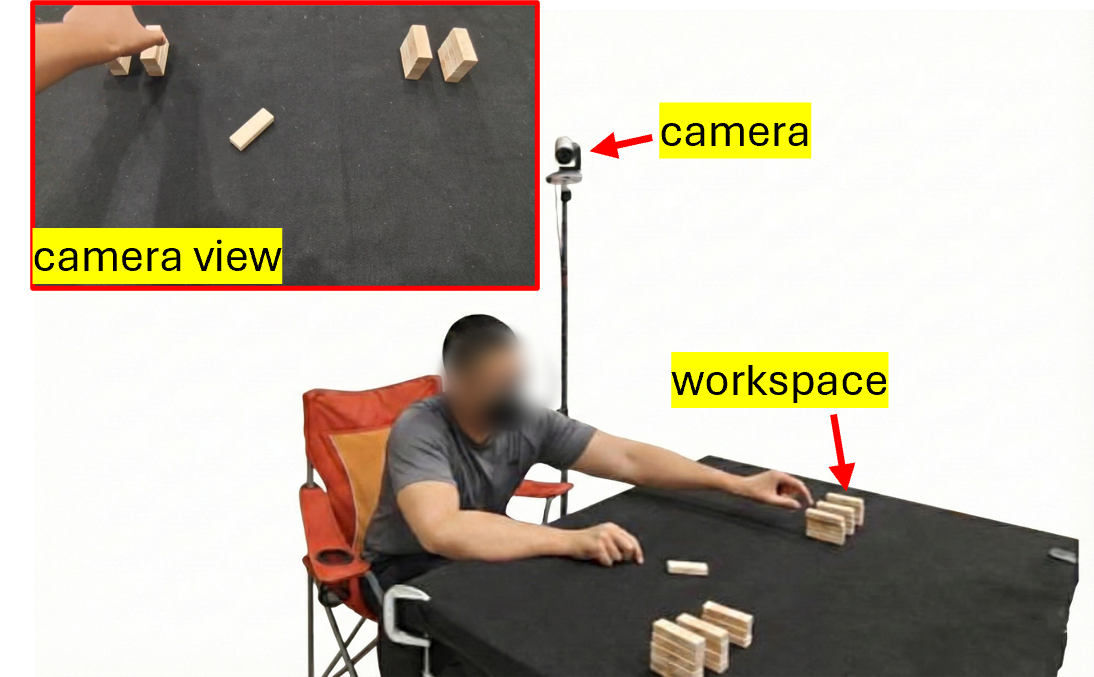}
        \subcaption{}
        \label{fig1::setup_sub_1} 
        \includegraphics[width=0.8\linewidth]{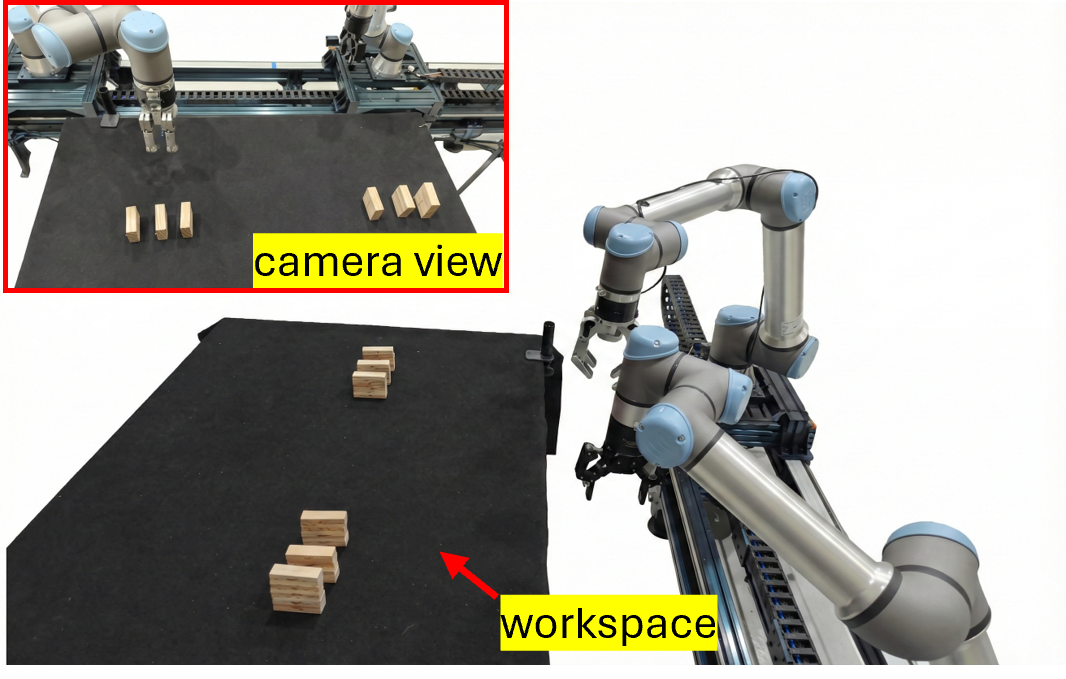}
        \subcaption{}
        \label{fig1::setup_sub_2} 
    \end{subfigure}
    \caption{(a) Configuration of the experimental environment and the associated camera viewpoint during the human demonstration phase. (b) The dual-arm robot system workspace and the visual feed are captured as the robot executes the task.}
    \label{fig::exp_setup}
\end{figure}

Our method was evaluated through a series of four experiments, each targeting a critical component of the proposed framework. The overall system architecture is visualized in Fig.~\ref{fig::workflow_all}, detailing how raw multimodal inputs are abstracted into symbolic scene graphs to support robust policy transfer. These structured representations serve as priors for the Planning Agent, which coordinates with the Action Head to drive the dual-arm system through a continuous perception-action loop.

(1) Video Processing and Task Representation:
The first experiment aimed to assess the functionality of the video processing module and the quality of the task representations derived from it. These representations were used as inputs to the large language model (LLM), serving as the foundation for downstream task planning and policy generation.

(2) Evaluation of LLM-Based Task Policy Generation and Reasoning:
The second experiment focused on evaluating the capabilities of the LLM head within the VLA· framework. Specifically, we assessed:
(i) its ability to generate an overall task policy that governs the robot's control strategy,
(ii) its competence in decomposing the high-level task into a sequence of coherent and executable subtasks,
(iii) its capacity to produce Chain-of-Thought (CoT) reasoning traces that explicitly articulate the rationale behind each subtask decision, and
(iv) its ability to perform self-verification, including both logical consistency checks of the generated plan and the capability to assess whether each executed subtask has been successfully and adequately completed.

(3) Assessment of Fine-Tuned VLA in Block Manipulation Tasks:
In the third experiment, we evaluated the low-level manipulation capabilities of the fine-tuned VLA system in a series of object-centric tasks involving the grasping and placement of blocks with varying shapes, sizes, and spatial configurations. This experiment aimed to assess the generalization performance of the model across diverse physical layouts and object combinations, as well as its ability to maintain stable execution across multiple trials. The evaluation focused on success rates of grasp and place actions, and precision in object alignment. 

(4) Execution Performance Based on LLM-Generated Policy:
We conducted an end-to-end evaluation to assess the system’s real-world performance under policies generated solely by the LLM head. Beyond measuring action accuracy and perception-to-action robustness, this experiment tested the model’s reasoning capability in planning complex multi-step tasks, such as constructing letter-shaped structures using varying numbers of blocks and coordinating dual-arm assembly sequences. Robustness was further evaluated under perturbations, including block misplacement and human-induced shifts. The system demonstrated the ability to detect inconsistencies, replan locally, and recover autonomously, highlighting its resilience and reasoning competence in dynamic execution scenarios.

\subsection{Task representation assessment}
The assessment dataset consists of 250 videos collected from the Subsection \ref{sec:training}. 
The tasks were designed to capture diverse modes of object interaction, including:
(1) Simple single-hand manipulation, where a single block was moved using one hand, to evaluate basic hand–object interaction dynamics; and
(2) Multi-block assembly, in which participants constructed meaningful structures using multiple blocks, to analyze spatial relationships and inter-object coordination during sequential manipulation.
The image plane was aligned parallel to the working surface to ensure accurate spatial correspondence between the captured image and the physical task space.
To enable robust detection and segmentation of both hands and objects, including 6D pose estimation, we employed the real-time \textit{SAM2} segmentation framework. As the tasks mainly involved planar motions, computations of spatial relationships and entropy were based on the $x$ and $y$ components of the position signals. Nevertheless, for completeness, each graph node retained full 3D position and orientation information.
RGB data were captured at 30~Hz. A temporal window of $\phi = 1\,\text{s}$ (40 frames) was used to compute average distances and entropy measures. For each time point $t$, hand and object positions were analyzed within a symmetric window spanning approximately 0.5 seconds before and after $t$, which aligns with the typical time required for participants to alter object relationships.
Positional data were quantized at a resolution of $\xi = 1\,\text{cm}$ for entropy calculations. We set key thresholds as follows: minimum mutual information $\varepsilon_{\mathrm{MI}} = 0.05$, hand--object interaction distance $d_{\mathrm{th}} = 0.15\,\text{m}$, and object--object proximity $d_{\mathrm{oo}} = 0.2\,\text{m}$, based on marker-based localization. These parameters were empirically optimized through extensive preliminary validation across the dataset to maximize interaction detection accuracy, and the system demonstrates robust performance stability within a $\pm 10\%$ margin of these selected thresholds.
To evaluate signal dynamics, we examined the most recent 20 samples ($w/2$) at each time point. A signal was considered to be decreasing if the majority of consecutive differences showed a downward trend, indicating a reduction in spatial proximity or activity.

We evaluate the quality of the video-to-representation pipeline using two core metrics:

\paragraph{Graph Representation Accuracy (GRA)}
This metric quantifies the correctness of the extracted interaction graph $\mathcal{G}_t$ at each time step, compared to a ground-truth graph $\mathcal{G}^*_t$ annotated from labeled keyframes.

\begin{equation}
\mathrm{GRA} = \frac{1}{T} \sum_{t=1}^T \mathbb{I}[\mathcal{G}_t = \mathcal{G}^*_t]
\end{equation}

where $\mathbb{I}[\cdot]$ is the indicator function and $T$ is the number of evaluated time steps (or frames). Two graphs are considered equal if their nodes and edges match in both topology and associated object identities within a spatial tolerance $\delta_{\text{pos}} < 2\,$cm.
GRA reflects the system's ability to correctly extract the spatial-temporal structure of the task (i.e., which objects are involved and how they interact) from RGB-D data.

\vspace{1em}

\paragraph{Task Segmentation Accuracy (TSA)}
This metric evaluates how accurately the system segments a continuous video sequence into meaningful sub-tasks (e.g., individual block placements) by comparing predicted segmentation boundaries $\hat{B} = \{\hat{t}_1, \hat{t}_2, \dots\}$ with ground-truth boundaries $B^* = \{t^*_1, t^*_2, \dots\}$.

We adopt a boundary-matching metric adapted from action segmentation literature:

\begin{equation}
\mathrm{TSA} = \frac{1}{|B^*|} \sum_{i=1}^{|B^*|} \mathbb{I}\left[ \min_j |\hat{t}_j - t^*_i| \leq \tau \right]
\end{equation}

where $\tau$ is a temporal tolerance threshold (set to $0.5$ seconds), and $|B^*|$ is the total number of ground-truth subtask boundaries.
TSA measures whether each subtask transition (e.g., placing a new block) is correctly identified by the model within a permissible time window.

\subsection{Task Planning Assessment}
\label{sec:task_planning_assessment}

To quantitatively assess the performance of the VLA’s LLM head in generating human-like task policies and reasoning chains, we conducted an evaluation using 125 RGB video demonstrations that were held out specifically for testing. These videos were collected from ten human participants and are distinct from the 125 samples used during training. For each testing video, a ground-truth policy was constructed by domain experts through manual annotation, including structured decomposition of subtasks, their temporal ordering, and inferred task intentions.
The LLM-generated outputs were evaluated against these reference annotations along four dimensions:

\paragraph{Plan Coverage ($PC$)}
This metric measures the proportion of ground-truth subtasks that were successfully identified by the LLM. Let $\mathcal{T}_{\text{gt}}$ denote the set of ground-truth subtasks and $\mathcal{T}_{\text{llm}}$ the set of LLM-generated subtasks. The coverage is computed as:
\begin{equation}
PC = \frac{|\mathcal{T}_{\text{gt}} \cap \mathcal{T}_{\text{llm}}|}{|\mathcal{T}_{\text{gt}}|}
\end{equation}
Higher values indicate better subtask recognition.

\paragraph{Ordering Accuracy ($OA$)}
This metric evaluates whether the temporal sequence of subtasks generated by the LLM matches that of the human demonstration. We compute the normalized Kendall’s Tau correlation coefficient:
\begin{equation}
OA = \frac{1 + \tau(\mathcal{T}_{\text{gt}}, \mathcal{T}_{\text{llm}})}{2}
\end{equation}
where $\tau(\cdot, \cdot)$ denotes Kendall’s Tau. The value is normalized to fall within $[0, 1]$, with $OA = 1$ indicating perfect temporal alignment.

\paragraph{Chain-of-Thought Interpretability ($CI$)}
CoT reasoning traces were rated by three domain experts on a 5-point Likert scale (1 = incoherent, 5 = fully interpretable and goal-consistent). The final interpretability score is:
\begin{equation}
CI = \frac{1}{3N} \sum_{i=1}^{N} \sum_{j=1}^{3} r_{ij}
\end{equation}
where $r_{ij}$ is the score given by expert $j$ for trial $i$, and $N$ is the total number of evaluated trials.

\paragraph{Verification Correctness ($VC$)}
This metric evaluates whether the LLM correctly determined the success or failure of each subtask. Let $v_i$ be the ground-truth success label and $\hat{v}_i$ the LLM’s prediction for subtask $i$. Then:
\begin{equation}
VC = \frac{1}{|\mathcal{T}_{\text{gt}}|} \sum_{i=1}^{|\mathcal{T}_{\text{gt}}|} \mathbb{1}(v_i = \hat{v}_i)
\end{equation}
where $\mathbb{1}(\cdot)$ is the indicator function. Higher values indicate better self-assessment capability.

\subsection{Block manipulation assessment}
\label{sec:task_execution_assessment}

In the third experiment, we evaluated the low-level manipulation performance and generalization ability of the fine-tuned GF-VLA model across a series of dual-arm block-based tasks involving variations in object shape, spatial relations, and instruction types. The goal was to assess the system’s ability to execute robust grasping, maintain coordinated dual-arm motion, and achieve precise object placement under diverse shape, pose, and semantic instruction conditions.
This evaluation utilized the held-out portion of the dataset described in Section~\ref{sec:training}, specifically the remaining 120 trials not used during Action Head fine-tuning. These consisted of four task categories (shape generalization, spatial relation grounding, absolute 6D pose execution, and relative 6D pose execution), each performed with three distinct block shapes (cuboid, triangular prism, and cube), with 10 repetitions per shape per task. This resulted in $4$ tasks $\times$ $3$ shapes $\times$ $10$ trials = 120 trials in total for quantitative evaluation.

For each execution, we recorded the following performance metrics:

\paragraph{Grasp Success Rate (GSR)}
The proportion of successful grasp attempts over all trials is defined as
$\mathrm{GSR} = \frac{N_{\text{grasp\_success}}}{N_{\text{grasp\_attempts}}}$.

\paragraph{Place Success Rate (PSR)}
The proportion of successful placements (i.e., stable and within tolerance) is computed as
$\mathrm{PSR} = \frac{N_{\text{place\_success}}}{N_{\text{place\_attempts}}}$.

\paragraph{6D Pose Placement Error (6DPE)}
This metric evaluates the placement accuracy in both position and orientation:
\begin{equation}
\mathrm{6DPE} = \sqrt{||\mathbf{p}\text{est} - \mathbf{p}\text{gt}||^2} + \lambda \cdot \theta
\end{equation}
where $\mathbf{p}\text{est}$ and $\mathbf{p}\text{gt}$ are the estimated and ground-truth 3D positions, $\theta$ is the angular deviation in radians, and $\lambda$ is a balancing coefficient.

\paragraph{Instruction Compliance Score (ICS)}
This metric reflects how well the robot's behavior complies with spatial instructions, especially ambiguous ones. Experts rated each trial using a 5-point Likert scale. The final score is calculated as:
\begin{equation}
\mathrm{ICS} = \frac{1}{kN} \sum_{i=1}^{N} \sum_{j=1}^{k} r_{ij}
\end{equation}
where $r_{ij}$ is the rating from expert $j$ for trial $i$, $k$ is the number of experts, and $N$ is the number of evaluated trials.

\medskip

This experiment enabled a comprehensive analysis of the GF-VLA model’s capabilities in low-level motion execution, spatial generalization, semantic instruction following, and coordinated dual-arm control. By combining physical shape variations with both ambiguous and precise spatial instructions, and evaluating performance under structured trial conditions, we validated the system’s ability to generalize manipulation skills across a wide range of real-world block assembly scenarios.

\subsection{Task manipulation assessment}

This experiment evaluates the generalization capability of our proposed framework in transferring policies generated from human demonstrations to novel but structurally related tasks. 
Specifically, we assess whether a dual-arm robot system can successfully perform cooperative block assembly tasks—constructing symbolic letter shapes and geometric towers—under varying initial conditions and target configurations.

We focused on two categories of dual-arm collaborative manipulation tasks: Letter Assembly Tasks, involving the construction of composite symbolic structures (e.g., “VLM”) using rectangular and square blocks arranged in predefined configurations; and
Tower Construction Tasks, where both arms cooperatively assembled towers of various geometric profiles (e.g., triangular base towers, vertical square towers), composed from identical unit blocks.

Each task was first demonstrated by a human using both hands. In particular, the demonstration of constructing the letter “R” was recorded and converted into an information-theoretic behavior graph, which served as input to the GF-VLA model. The model then generated a task plan, subtask sequence, and corresponding chain-of-thought reasoning for execution. This policy, derived from a single human demonstration, was subsequently applied to a series of modified scenarios to evaluate generalization performance. The dual-arm robot executed the inferred policy under three key variations: (i) constructing new composite letter targets such as “VLM”, which introduced increased spatial complexity and extended block arrangements; (ii) performing under altered block configurations and randomized initial poses; and (iii) adapting to different workspace conditions, including shifted camera viewpoints and the removal of irrelevant distractor objects. In addition to symbolic letter tasks, the same policy trained on a triangular-base tower was reused to assemble alternative tower configurations, such as vertical square towers and asymmetric block stacks, with both arms executing in coordinated motion. Each variation was repeated 10 times, yielding a total of 60 trials (3 letter assemblies and 3 tower configurations).

To further validate the efficacy of fusing structured information-theoretic priors with large language models, we compared the proposed GF-VLA framework against five leading baseline methods—GraspVLA~\cite{deng2025graspvla}, Openhelix~\cite{cui2025openhelix}, GR00T N1~\cite{bjorck2025gr00t}, ChatVLA~\cite{zhou2025chatvla}, and DiffusionVLA~\cite{wen2025diffusionvla}—across the two most challenging generalization scenarios: the ``Building Letter`` assembly and the ``Building Tower" construction.

The generalization performance was evaluated using the following three metrics, each quantifying a different aspect of dual-arm execution robustness and policy adaptability.

\paragraph{Task Success Rate (TSR)}
TSR quantifies the proportion of trials in which the final structure matched the intended configuration (letter or tower). It reflects the reliability of the execution pipeline from policy generation to motion completion:
$\mathrm{TSR} = \frac{N_{\text{success}}}{N_{\text{total}}}$
where $N_{\text{success}}$ is the number of successful executions and $N_{\text{total}}$ is the total number of attempted trials.

\paragraph{Bimanual Coordination Score (BCS)}
BCS captures the fluency and synchrony of the two robotic arms during task execution. Each trial was scored by expert evaluators on a 5-point Likert scale based on the smoothness, timing, and collaborative efficiency of the bimanual motions. The final score is computed as:
\begin{equation}
\mathrm{BCS} = \frac{1}{N} \sum_{i=1}^{N} s_i
\label{eq:bcs}
\end{equation}
where $s_i$ is the score assigned to trial $i$ and $N$ is the total number of trials.

\paragraph{Plan Transferability Rate (PTR)}
PTR measures the effectiveness of reusing a policy derived from a single human demonstration to accomplish novel task variants without retraining. It indicates the system’s capacity for generalization:
$\mathrm{PTR} = \frac{N_{\text{reused\_success}}}{N_{\text{reused\_attempts}}}$
where $N_{\text{reused\_success}}$ is the number of successfully completed transfer executions, and $N_{\text{reused\_attempts}}$ is the total number of such attempts.

\section{Results and analysis}

\subsection{Task representation assessment}
\label{subsec:exp1_video}

\begin{figure*}[!t]
    \centering
    \includegraphics[width=1\linewidth]{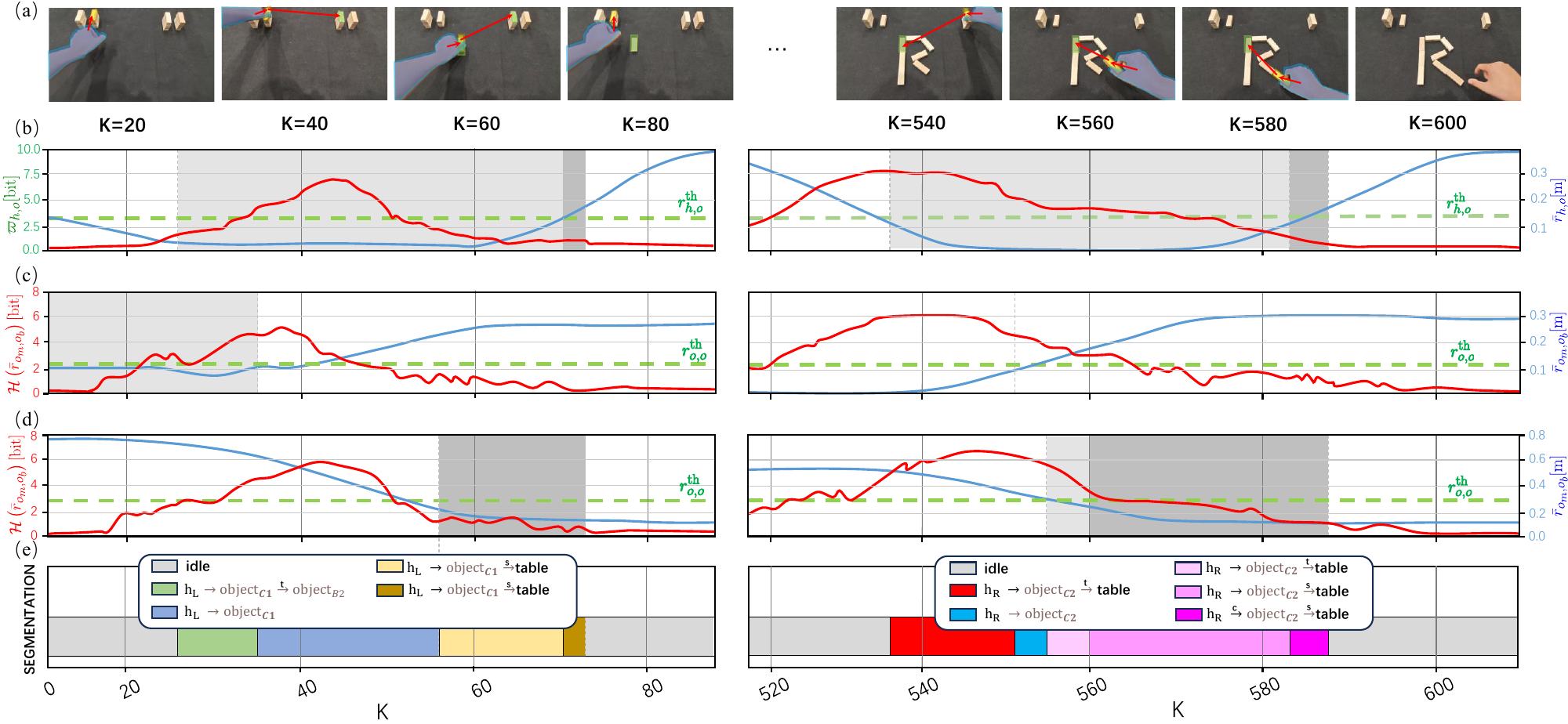}
    \caption{Information-theoretic analysis and temporal segmentation of a human demonstration for a block assembly task. (a) Keyframes from the demonstration video showing the human assembling the letter ``R". (b) Analysis of the Hand-Object (HO) interaction, plotting the mutual information ($\boldsymbol{\varpi}_{h,o}$) and relative distance ($\overline{r}_{h,o}$) between the hand and the manipulated block to detect kinematic coupling. (c)-(d) Analysis of Object-Object (OO) interactions, plotting the entropy of the relative distance ($\mathcal{H}(\overline{r}_{o_m,o_b})$) and the distance itself ($\overline{r}_{o_m,o_b}$) between the in-hand block and other objects to identify stable placements. (e) The resulting temporal segmentation of the task classifies the demonstration into a sequence of distinct interaction primitives based on the information-theoretic metrics.}
    \label{fig:demon_R_res}
\end{figure*}

The first experiment aimed to assess the functionality of the video processing module and the quality of the task representations derived from it. These representations served as input to the LLM, forming the foundation for downstream task planning and policy generation.

% Figure~\ref{fig:demon_R_res} illustrates a representative scene graph extracted from a human demonstration of constructing the letter "R" with blocks. Figure~\ref{fig:demon_R_res}(a) shows how SAM2 model segment hand and jenga block using keyframes to summarize the scene evolution from approach and grasp the jenga blocks to transport, alignment, placement, and retreat from $K=0$ to $K= 800$. Figure~\ref{fig:demon_R_res}(b) shows

In the task that building letter ``R"  by dual hand as illustrated in Fig.~\ref{fig:demon_R_res}, the participant undertook to choose which hand first(left hand $h_L$ or right hand $h_R$) and then subsequently place one $\mathrm{object_{C1}}$ using one of hands onto the workspace, within a context containing two groups of visible objects. Throughout this manipulation, $\mathrm{object_{C1}}$ momentarily maintained a $OO$ interaction with $object_{C1}$ positioned on the table, due to their proximity measure $\bar{r}_{o,o}$ remaining below the established threshold at timestep $k=40$. Nevertheless, this interaction was categorized as \texttt{T-OO} owing to an observed increase in entropy $\mathcal{H}(\bar{r}_{o,o})$. Such entropy behavior implies that explicitly encoding the interaction with $\mathrm{table}$ might be unnecessary for accurately completing the task. Alternative demonstrations with minor variations in the initial position of $\mathrm{object_{C1}}$ revealed a different handling strategy. In these cases, the hand's trajectory typically navigated along the box edge, perceiving it as an obstacle requiring avoidance before reaching the workspace. Consequently, the corresponding $OO$ interaction in these instances was regarded as \texttt{E-OO}. This significant interaction persisted until sufficient separation occurred between the involved objects, a condition visualized by the shaded blue region at the left side of Fig.~\ref{fig:demon_R_res}(c). This separation was marked explicitly at timestep $k=80$, coinciding with a recorded modification in the scene graph topology. Approaching the table, another \texttt{E-OO} relationship developed between $\mathrm{object_{C1}}$ and the workspace, formally noted at timestep $k=70$. This interaction was identified through the decreasing trend in entropy $\mathcal{H}(\bar{r}_{o,o})$, as depicted by the descending trajectory of the red curve in Fig.~\ref{fig:demon_R_res}(d). Notably, no scene graph was produced during the weighing phase since the hand disengaged from $\mathrm{object_{C1}}$. Upon re-grasping $\mathrm{object_{C2}}$, a brief \texttt{T-OO} emerged with the workspace due to initial close proximity, but this relation was rapidly dissolved, reflected by a swift rise in the entropy trend $\mathcal{H}(\bar{r}_{oo})$. The corresponding scene graph configuration at $K=552$ remained stable until $\mathrm{object_{C2}}$ eventually distanced itself sufficiently from the workspace. Further interaction was registered when $\mathrm{object_{C2}}$ already placed on the table by right hand has been encountered. The \texttt{T-OO} transitioned to \texttt{E-OO} as soon as entropy $\mathcal{H}(\bar{r}_{oo})$ began decreasing between the two object. The commencement of this transition corresponds with the darker grey highlighted region in Fig.~\ref{fig:demon_R_res}(d). Nonetheless, the overall topology of the scene graph remained unchanged from timesteps $K=560$ to $K=585$. The evolution and changes within the scene graph are clearly visualized by the colored bars at the lower portion of Fig.~\ref{fig:demon_R_res} (e).

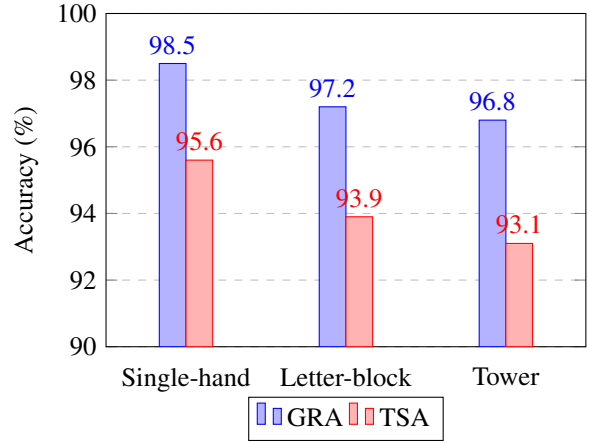
\begin{figure}[!t]
\centering
\begin{tikzpicture}
\begin{axis}[
    ybar=0pt,
    bar width=10pt,
    width=0.48\textwidth,
    height=6cm,
    ymin=90,
    ymax=100,
    enlarge x limits=0.25,
    ylabel={Accuracy (\%)},
    symbolic x coords={Single-hand, Letter-block, Tower},
    xtick=data,
    nodes near coords,
    legend style={at={(0.5,-0.15)}, anchor=north, legend columns=-1},
    ylabel near ticks,
    xtick style={draw=none},
    ymajorgrids=true,
    grid style=dashed
]
% GRA: Graph Representation Accuracy
\addplot+[style={blue,fill=blue!30}] coordinates {
    (Single-hand,98.5)
    (Letter-block,97.2)
    (Tower,96.8)
};
% TSA: Task Segmentation Accuracy
\addplot+[style={red,fill=red!30}] coordinates {
    (Single-hand,95.6)
    (Letter-block,93.9)
    (Tower,93.1)
};
\legend{GRA, TSA}
\end{axis}
\end{tikzpicture}
\caption{Comparison of Graph Representation Accuracy (GRA) and Task Segmentation Accuracy (TSA) across different task types.}
\label{fig:gra_tsa_bar}
\end{figure}

The results shown in Figure~\ref{fig:gra_tsa_bar} demonstrate that our method achieves consistently high performance across all task types for both Graph Representation Accuracy (GRA) and Task Segmentation Accuracy (TSA). In the single-hand manipulation task, which involves minimal interaction complexity and well-separated actions, the system attained the highest GRA of 98.5\% and TSA of 95.6\%. This reflects the clarity of visual signals and subtask transitions in scenarios involving isolated object movement. In the letter-block assembly task, although visual complexity increased due to bimanual coordination and densely arranged blocks, the GRA remained high at 97.2\%, indicating the robustness of spatial graph extraction even in multi-object interactions. The TSA in this task was slightly lower (93.9\%), likely due to temporal overlaps between subtasks and occasional ambiguity in the onset of block placement actions. The tower construction task posed additional challenges such as vertical stacking, depth occlusion, and faster-paced execution. While GRA stayed strong at 96.8\%, the TSA dropped modestly to 93.1\%, suggesting that temporal segmentation is more sensitive to rapid task transitions and occlusions than graph structure inference. Overall, these results confirm that our framework effectively captures both spatial and temporal structures in diverse manipulation tasks, with performance degrading gracefully under increased task complexity.

\subsection{Task planning assessment}

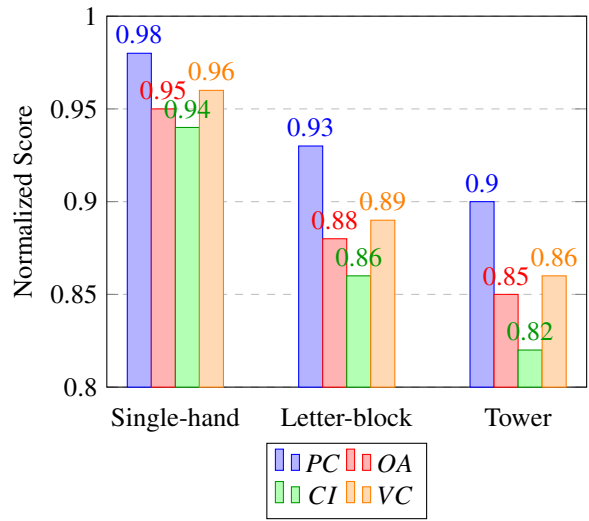
\begin{figure}[!t]
\centering
\begin{tikzpicture}
\begin{axis}[
    ybar=0pt,
    bar width=9pt,
    width=0.48\textwidth,
    height=6.5cm,
    ymin=0.8,
    ymax=1.0,
    enlarge x limits=0.2,
    ylabel={Normalized Score},
    symbolic x coords={Single-hand, Letter-block, Tower},
    xtick=data,
    nodes near coords,
    legend style={at={(0.5,-0.15)}, anchor=north, legend columns=2},
    xtick style={draw=none},
    ymajorgrids=true,
    grid style=dashed
]
\addplot+[style={blue,fill=blue!30}] coordinates {
    (Single-hand,0.98)
    (Letter-block,0.93)
    (Tower,0.90)
};
\addplot+[style={red,fill=red!30}] coordinates {
    (Single-hand,0.95)
    (Letter-block,0.88)
    (Tower,0.85)
};
\addplot+[style={green!60!black,fill=green!30}] coordinates {
    (Single-hand,0.94)
    (Letter-block,0.86)
    (Tower,0.82)
};
\addplot+[style={orange,fill=orange!30}] coordinates {
    (Single-hand,0.96)
    (Letter-block,0.89)
    (Tower,0.86)
};
\legend{$PC$, $OA$, $CI$, $VC$}
\end{axis}
\end{tikzpicture}
\caption{LLM-generated plan and reasoning evaluation scores across tasks. All scores normalized to [0, 1] for visualization.}
\label{fig:vla_llm_bar}
\end{figure}

The results in Figure~\ref{fig:vla_llm_bar} demonstrate that the VLA model's LLM head consistently produces high-quality task plans and reasoning traces across different task categories. For single-hand manipulation, the system achieved near-perfect performance with 98\% plan coverage, 95\% ordering accuracy, and a verification correctness of 96\%, reflecting the relative simplicity and linearity of the action sequences. The chain-of-thought (CoT) explanations were rated highly by human experts, with an average interpretability score of 4.7 out of 5.

In the more structurally complex letter-block assembly tasks, the LLM still maintained strong performance, with 93\% subtask coverage and 88\% ordering alignment. Minor ordering deviations often occurred in symmetric or parallel subtasks (e.g., placing the left and right legs of the letter “A” in reverse), which had no effect on overall task validity. CoT reasoning remained coherent (4.3/5), though some rationales became verbose or under-specified. Verification correctness reached 89\%, indicating good—but not perfect—self-evaluation capacity.

For tower construction, challenges included overlapping subgoal transitions, height occlusion, and ambiguous stacking order. These factors led to slightly reduced coverage (90\%) and ordering accuracy (85\%), though still within an acceptable range. CoT interpretability remained relatively high (4.1), but experts noted occasional overgeneralization (e.g., “stacking the top” without specifying which block). Verification accuracy (86\%) reflects occasional failure to detect partially misaligned blocks.

Overall, the LLM head shows strong capacity to infer subgoal structure, temporal dependencies, and causal rationales across diverse manipulation tasks. The CoT traces not only improve transparency but also support downstream verification and adaptive correction. The results also highlight areas where improvements—particularly in fine-grained temporal modeling and object-level reasoning—could further enhance generalization and robustness.

\subsection{Block manipulation assessment}

\begin{table*}[t]
\centering
\caption{Performance of Dual-Arm Manipulation under Different Task Conditions}
\begin{tabular}{lcccc}
\toprule
\textbf{Task Type} & \textbf{GSR} & \textbf{PSR} & \textbf{6DPE (cm + deg)]} & \textbf{ICS} \\
\midrule
Shape Generalization           & 0.98 $\pm$ 0.02 & 0.95 $\pm$ 0.03 & 1.2 cm + 2.5° & 4.8 $\pm$ 0.3 \\
Spatial Relation (Ambiguous)   & 0.94 $\pm$ 0.03 & 0.90 $\pm$ 0.04 & 2.1 cm + 4.2° & 4.5 $\pm$ 0.5 \\
Absolute 6D Pose Execution     & 0.92 $\pm$ 0.04 & 0.87 $\pm$ 0.05 & 2.8 cm + 6.0° & 4.2 $\pm$ 0.6 \\
Relative Pose Execution        & 0.91 $\pm$ 0.04 & 0.85 $\pm$ 0.06 & 3.1 cm + 6.8° & 4.4 $\pm$ 0.4 \\
\midrule
\textbf{Overall Average}       & \textbf{0.94}   & \textbf{0.89}   & \textbf{2.3 cm + 4.9°} & \textbf{4.5} \\
\bottomrule
\end{tabular}
\label{tab:gfvla_manipulation_results}
\end{table*}

The results summarized in Table~\ref{tab:gfvla_manipulation_results} confirm the GF-VLA model's capacity to execute complex dual-arm manipulation tasks with high success rates and spatial precision across diverse conditions. In the shape generalization task, the system achieved the highest grasp success rate (98\%) and placement success rate (95\%), demonstrating reliable perception-to-action mapping even when dealing with unseen shapes. The average 6D pose placement error was minimal (1.2 cm and 2.5°), and the instruction compliance score reached 4.8 out of 5, indicating that the LLM head effectively grounded shape-specific references into motor actions.

For ambiguous spatial relation tasks, such as “place the long block next to the triangle,” the system maintained strong performance with a 94\% grasp and 90\% placement success rate. Despite the vagueness of the instruction, the GF-VLA model leveraged contextual cues and visual relations to interpret adjacency constraints, reflected in a high compliance score (4.5). The slightly higher 6D error (2.1 cm, 4.2°) suggests that spatial softness in instruction also leads to looser execution boundaries.

In the absolute 6D pose execution task, the system faced increased difficulty due to the precise spatial requirements and the need for accurate camera-to-arm calibration. Although the success rates declined slightly (92\% grasp, 87\% placement), the model still produced low average pose errors (2.8 cm + 6.0°), showing that it could generalize explicit goals into coordinated dual-arm control with acceptable precision. The instruction compliance score (4.2) remained high, reflecting consistent alignment with intended goals.

The relative pose task was the most challenging, involving both semantic interpretation and relational spatial reasoning. While grasp and placement success were slightly lower (91\% and 85\%, respectively), the compliance score of 4.4 demonstrates the model’s capacity to understand and realize geometric spatial relations across blocks (e.g., diagonal crossing, top-down intersection). The average pose error (3.1 cm, 6.8°) reflects cumulative uncertainties in relative frame estimation and execution.

Across all tasks, the model maintained strong dual-arm coordination despite differing gripper types and role assignments. The UR5e arm (Robotiq) typically handled source block retrieval, while the UR10e (Barrett Hand) contributed to stabilization, alignment, or relational framing. These results support the conclusion that GF-VLA not only interprets diverse spatial commands—including ambiguous and reference-relative ones—but also generalizes well to novel shapes and unseen block combinations under real-world physical constraints. 

\subsection{Albation Study}

To validate the specific contributions of each module within the Graph Fused VLA framework, we conducted a systematic ablation study with experiments in Subsection 5.4 as summarized in Table~\ref{tab:albation_study}. The most critical degradation in performance was observed when the information-theoretic scene graph was removed from the input stream. This variant relies solely on unstructured visual embeddings and resulted in an average decrease of 0.49 in TSR. The decline was particularly evident in the Letter Assembly task where precise spatial relations are paramount. This finding confirms that the entropy-based interaction primitives provide a crucial foundation for the planner, enabling the system to distinguish between incidental contact and meaningful manipulation. Without this structured prior the policy struggles to maintain kinematic consistency over long horizons and frequently fails to stabilize complex object configurations.

We further evaluated the impact of the CoT reasoning mechanism and the unified dual head architecture. Removing the explicit reasoning steps resulted in a 0.17 drop in TSR on average, indicating that the intermediate generation of subgoals and self-verification tokens is crucial for handling complex sequential dependencies in dual-arm tasks. When the model operates without the dual head design and uses a single output stream for both language and control, the performance decreases by 0.37. This reduction suggests that decoupling the semantic planning space from the continuous action space allows for more effective feature specialization during fine-tuning. Consequently, the full framework achieves the highest robustness by effectively synergizing structured perception with hierarchical reasoning.

\begin{table}[!htbp]
\centering

\caption{Ablation study quantifying the impact of key GF-VLA components on dual-arm manipulation TSR.}
\resizebox{0.9\columnwidth}{!}{% Resize to fit column if necessary
\begin{tabular}{lcccc}
\toprule
\textbf{Model Variant} & \textbf{Component Removed} & \textbf{\begin{tabular}[c]{@{}c@{}}Letter\\ Assembly\end{tabular}} & \textbf{\begin{tabular}[c]{@{}c@{}}Tower\\ Construction \end{tabular}} & \textbf{\begin{tabular}[c]{@{}c@{}}Average\\ Decrease\end{tabular}} \\ \midrule
\textbf{Full GF-VLA (Ours)} & \textit{None} & \textbf{0.88} & \textbf{0.92} & - \\
Var A: w/o Scene Graph & Info-Theoretic Graph & 0.48 & 0.35 & $\downarrow$ 0.49 \\
Var B: w/o CoT & Chain-of-Thought & 0.76 & 0.71 & $\downarrow$ 0.17 \\
Var C: w/o Dual-Head & Unified Dual-Head & 0.51 & 0.56 & $\downarrow$ 0.37 \\ \bottomrule
\end{tabular}}
\label{tab:albation_study}
\end{table}

\subsection{Task manipulation assessment}

\begin{table}[t]
\centering
\caption{Generalization Performance on Dual-Arm Policy Transfer Tasks}
\begin{tabular}{lccc}
\toprule
\textbf{Assembled Object} & \textbf{TSR} & \textbf{BCS} & \textbf{PTR} \\
\midrule
Letter: “VLM”             & 0.90 & 4.6 $\pm$ 0.4 & 0.85 \\
Letter:  randomized poses  & 0.87 & 4.4 $\pm$ 0.5 & 0.83 \\
Letter:  distractor present & 0.88 & 4.5 $\pm$ 0.5 & 0.84 \\
Tower: square tower                & 0.93 & 4.7 $\pm$ 0.3 & 0.89 \\
Tower: asymmetric stack            & 0.90 & 4.5 $\pm$ 0.4 & 0.87 \\
Tower: shifted viewpoint           & 0.92 & 4.6 $\pm$ 0.4 & 0.88 \\
\midrule
\textbf{Overall Average}           & \textbf{0.90} & \textbf{4.55} & \textbf{0.86} \\
\bottomrule
\end{tabular}
\label{tab:generalization_transfer}
\end{table}

The results in Table~\ref{tab:generalization_transfer} demonstrate that the GF-VLA framework successfully generalizes dual-arm task policies derived from a single human demonstration to structurally related yet novel manipulation scenarios. Across all six task variations, the overall task success rate (TSR) reached 90\%, confirming the robustness of the end-to-end system in executing composite goals under variation in object arrangement, pose, and perception conditions.

For the symbolic letter assembly tasks, which involve complex spatial configurations and semantic structure, TSR ranged between 87\% and 90\%. The slight drop under randomized initial poses suggests that the LLM-generated plan preserved high-level task intent but required additional low-level correction during execution. Despite this, the plan transferability rate (PTR) remained above 83\% across all three letter assembly variants, indicating that policies learned from the “R” demonstration generalized well even under workspace noise and distractors. Bimanual Coordination Scores (BCS) were consistently high ($\geq 4.4$ ), reflecting smooth, temporally aligned arm motions and effective division of labor between the Robotiq and Barrett grippers.

In tower construction tasks, where spatial relations are more geometric and less semantically constrained, the system achieved even higher TSR (up to 93\%) and PTR (up to 89\%). Notably, the shifted camera viewpoint had minimal impact on execution success, highlighting the model's spatial grounding resilience. The tower-based tasks also benefited from relatively symmetric motion primitives, which allowed more direct plan reuse. Coordination scores remained strong (BCS $\geq 4.5$ ), with evaluators noting the synchronized arm lifts and stable dual-arm placements.

\begin{table}[!htbp]
\centering
\caption{Comparative success rates of GF-VLA and state-of-the-art baselines on dual-arm "Building Letter VLM" and "Building Tower" tasks.}
\label{tab::3}
\resizebox{\linewidth}{!}{%\
% \color{red}{
\begin{tabular}{llcccccc}
\toprule
\textbf{Task} & \textbf{Metrics} & GraspVLA & Openhelix & GR00T N1 & ChatVLA & DiffusionVLA & \textbf{Ours} \\
\midrule
\multirow{1}{*}{Bulding Letter}
 & TSR                    & 0.79 & 0.83 & 0.81 & 0.87 & 0.85 & \textbf{0.88} \\
\midrule
\multirow{1}{*}{Building Tower}
 & TSR                    & 0.76 & 0.82 & 0.75 & 0.81 & 0.83 & \textbf{0.92} \\
\bottomrule
\end{tabular}}
\end{table}

The comparison with five leading baseline methods across the two most challenging generalization scenarios: the ``Building Letter" assembly and the ``Building Tower" construction is shown in Table~\ref{tab::3}. As indicated in the results, our method consistently outperforms the baselines, achieving a 0.88 TSR on the letter assembly task and 0.92 TSR on the tower construction task. This represents a significant improvement over purely generative approaches like ChatVLA and DiffusionVLA, which achieved 0.87 and 0.85 on the letter task, respectively. While methods such as ChatVLA leverage semantic reasoning, they often lack the explicit physical grounding provided by our method's scene graph representations, leading to occasional misalignments in complex spatial structures. The performance gap is even more pronounced when compared to execution-centric baselines like GraspVLA and GR00T N1, which trailed by margins of up to 0.17. This disparity can be attributed to the specific architectural advantages of GF-VLA. In the ``Building Letter" task, which requires high-level semantic planning and spatial reasoning, our integration of CoT prompting enables superior subtask decomposition compared to standard policy learners. Similarly, in the ``Building Tower" task, where geometric stability is paramount, the use of Shannon entropy and mutual information to quantify interaction dynamics enables our system to detect stable placement configurations more reliably than the diffusion-based or heuristic baselines. These comparisons empirically demonstrate that endowing VLA models with structured, information-theoretic inductive biases significantly enhances robustness and generalization in dual-arm manipulation tasks compared to relying solely on large-scale pretraining or end-to-end imitation.

Overall, these findings confirm the proposed system’s ability to generalize manipulation policies across task categories and environmental conditions with minimal supervision or retraining. The reuse of a single human demonstration to support six novel task variants—achieving $\geq 0.85$  success and strong coordination—demonstrates the effectiveness of the behavior graph representation and the LLM’s abstract reasoning capabilities in policy transfer.

\section{Discussion}

The experimental results across four evaluation axes collectively demonstrate the effectiveness, robustness, and generalization capability of the proposed GF-VLA framework in dual-arm robotic manipulation tasks. By integrating video-based perceptual representations, language-guided symbolic planning, and low-level action execution within a unified architecture, the system consistently achieved high task performance under diverse spatial, semantic, and environmental conditions.

First, in Experiment I, the video processing module successfully extracted temporally consistent spatial graphs from RGB-D streams, achieving over 95\% accuracy in graph representation and over 93\% subtask segmentation accuracy. These behavior graphs served as a crucial interface between raw perception and the LLM head, enabling symbolic abstraction of manipulation dynamics. The temporal entropy analysis and UMAP projections further confirmed the graph representations' stability and task-relevance across participants and conditions.

Experiment II assessed the quality of plans and reasoning chains generated by the LLM head. The model exhibited strong alignment with expert-labeled ground-truth in terms of plan coverage, temporal ordering, and self-verification. High chain-of-thought interpretability scores (avg. 4.4/5) suggest that the system's reasoning traces were not only functionally correct but also transparent and human-readable—an important step toward interpretable embodied intelligence. The performance remained stable across both symbolic and geometric task domains, demonstrating the model’s ability to ground semantic concepts into action structures.

Experiment III evaluated the low-level manipulation performance of the fine-tuned GF-VLA model across four bimanual task categories. The system maintained a 94\% grasp success rate and 89\% placement accuracy across shape, pose, and instruction variations. Even in ambiguous or relative spatial instruction scenarios, the model preserved high instruction compliance (avg. 4.5/5), reflecting its ability to interpret under-specified commands via visual and contextual grounding. The dual-arm coordination between UR5e and UR10e—equipped with heterogeneous grippers—remained stable across all trials, suggesting effective policy decomposition and motor-level role allocation.

Finally, Experiment IV demonstrated strong generalization from single-demonstration policies to novel task variants, achieving 90\% overall task success and 86\% policy transferability across both letter assembly and tower construction scenarios. This suggests that the combination of information-theoretic behavior graphs and LLM-based planning enables zero-shot task reuse under structural variations. The model adapted to shifted viewpoints, randomized configurations, and distractor interference without additional retraining, highlighting its robustness in realistic deployment conditions.

Despite these encouraging results, certain limitations remain. The segmentation of complex subtasks occasionally suffered under visually ambiguous or highly dynamic interactions, and plan ordering accuracy dropped in tasks involving parallel or symmetric subtasks. Moreover, the execution pipeline still relies on pre-calibrated camera-arm alignment and static workspace assumptions. Future work will explore end-to-end trainable modules for behavior graph extraction, incorporate temporal memory to handle task history, and extend the approach to more deformable or articulated objects. Additionally, incorporating self-correction or active failure recovery based on verification feedback could further enhance execution reliability.

In summary, GF-VLA establishes a modular yet tightly integrated framework that bridges perception, language, and action, enabling generalized, interpretable, and physically grounded dual-arm robotic behavior from minimal demonstration. Its demonstrated generalization across task forms, objects, and language conditions suggests promising applicability to real-world collaborative manipulation settings.

\section{Conclusion}
This work introduced GF‑VLA, a graph‑fused Vision Language Action framework that elevates Learning from Demonstration from motion imitation to task‑level policy reasoning. 
The method extracts Shannon Information cues to isolate the most informative hands–objects, encodes them as temporally ordered $HO$ and $OO$ interaction graphs, and fuses these with a language‑conditioned transformer to yield behavior trees and Cartesian commands; a cross‑hand selection policy further resolves bimanual role assignment without explicit geometry. Together with CoT reasoning and self‑verification in a dual‑head VLA design, the system delivers interpretable plans and executable actions from a single human video.

The four complementary experiments, which yielded related results, fully demonstrate the feasibility of the proposed approach across the full perception–planning–execution pipeline. They demonstrate that information-theoretic scene graphs yield stable, task-relevant abstractions; the language-guided policy head produces interpretable, expert-aligned plans; the integrated system executes robustly across variations in shape, pose, and instructions; and the learned policies generalize from a single demonstration to novel assemblies and viewing conditions. Together, these results indicate a reliable and transferable framework for observational learning in dexterous manipulation. The GF-VLA converted single human videos into executable dual-arm plans that succeeded with a high probability of block-stacking, letter-building, and tower-reconfiguration trials, while halving the 6D pose error relative to imitation-learning and language-only baselines, highlighting the value of fusing symbolic interaction structure with VLA reasoning.

Despite these advances, several limitations remain. Temporal segmentation can degrade in visually ambiguous or highly dynamic contacts, and subtask ordering accuracy dips for parallel/symmetric steps; moreover, execution still assumes pre‑calibrated camera–arm alignment and a largely static workspace. Future work will pursue end‑to‑end trainable graph extraction, temporal memory for long‑horizon context, and extension to deformable/articulated objects, alongside self‑correction and failure recovery driven by verification feedback.

% \section{Acknowledgment}
% This work was supported in part by the Natural Science Key Foundation of Zhejiang Province under Grant LZ24F030010, in part by the National Key Research and Development Program of China under Grant 
% 2021YFE01001\\00 and in part by the National Natural Science Foundation of China under Grant U1909209. We also sincerely thank the hardware support for experimental validations from the AgMan Lab at the University of New Mexico, which is funded by AFRL. 

\bibliographystyle{elsarticle-num} 
\bibliography{reference}
\end{document}